\documentclass[journal]{IEEEtran}
\usepackage{lmodern}
\usepackage{amsmath,amsthm,amssymb,mathtools,amsfonts}
\usepackage{graphicx}
\usepackage{enumitem,hyphenat,float,threeparttable}
\usepackage{textcomp}
\usepackage{xcolor}
\usepackage{algorithm}
\usepackage{algpseudocode}
\usepackage{xspace}
\usepackage{multirow}
\usepackage{booktabs}
\usepackage{thmtools}
\usepackage{hyperref}
\usepackage{zref-clever}

\usepackage[most]{tcolorbox}
\newtcbox{\mybox}[1][]{nobeforeafter,tcbox raise base,colframe=green!50!black,colback=green!10!white,top=0pt,bottom=0pt,left=0pt,right=0pt,before upper=\strut,#1}
\newtcbox{\myboxblue}[1][]{nobeforeafter,tcbox raise base,colframe=blue!50!black,colback=blue!10!white,top=0pt,bottom=0pt,left=0pt,right=0pt,before upper=\strut,#1}

\usepackage{tikz}
\usetikzlibrary{calc,shapes,arrows,positioning}
\usetikzlibrary{shapes.geometric}
\usetikzlibrary{arrows.meta}

\newcommand{\cafe}{\texttt{CAFe}\xspace}
\newcommand{\cafes}{\texttt{CAFe-S}\xspace}
\newcommand{\dcgd}{\texttt{DCGD}\xspace}
\newcommand{\dgd}{\texttt{DGD}\xspace}
\newcommand{\gd}{\texttt{GD}\xspace}
\newcommand{\cafelong}{\textbf{C}ompressed \textbf{A}ggregate \textbf{Fe}edback (\texttt{CAFe})\xspace}
\newcommand{\cafeslong}{\textbf{S}erver-Guided \textbf{C}ompressed \textbf{A}ggregate \textbf{Fe}edback (\texttt{CAFe-S})\xspace}
\newcommand{\dcgdlong}{\textbf{D}istributed \textbf{C}ompressed \textbf{G}radient \textbf{D}escent (\texttt{DCGD})\xspace}
\newcommand{\dgdlong}{\textbf{D}istributed \textbf{G}radient \textbf{D}escent (\texttt{DGD})\xspace}
\newcommand{\gdlong}{\textbf{G}radient \textbf{D}escent (\texttt{GD})\xspace}

\newcommand{\norm}[1]{\left\lVert#1\right\rVert}
\newcommand{\sqn}[1]{{\left\lVert#1\right\rVert}^2}
\newcommand\ec[2][]{\ensuremath{\mathbb{E}_{#1} \left[#2\right]}}

\newcommand\ev[1]{\left \langle{#1}\right \rangle}

\newcommand\br[1]{\left({#1}\right)}

\newcommand{\R}{\mathbb{R}}
\newcommand{\compress}[1]{\mathcal{C}\br{#1}}

\declaretheorem[name=Lemma]{lemma}
\declaretheorem[name=Definition, sibling=lemma]{definition}
\declaretheorem[name=Remark, sibling=lemma]{remark}
\declaretheorem[name=Assumption, style=plain, refname={assumption,assumptions}, Refname={Assumption,Assumptions}]{assumption}

\declaretheorem[name=Example]{example}

\newcommand{\cref}[1]{\zcref{#1}}
\newcommand{\Cref}[1]{\zcref[S]{#1}}

\zcRefTypeSetup{assumption}{
  name-sg = assumption,
  Name-sg = Assumption,
  name-pl = assumptions,
  Name-pl = Assumptions
}

\zcRefTypeSetup{figure}{
  name-sg = fig.,
  Name-sg = Fig.,
  name-pl = figs.,
  Name-pl = Figs.
}

\zcRefTypeSetup{equation}{
  name-sg = eq.,
  Name-sg = Eq.,
  name-pl = eqs.,
  Name-pl = Eqs.
}

\zcRefTypeSetup{algorithm}{
  name-sg = algorithm,
  Name-sg = Algorithm,
  name-pl = algorithms,
  Name-pl = Algorithms
}

\title{Communication Compression for Distributed Learning with Aggregate and Server-Guided Feedback}

\author{Tomas Ortega,
Chun-Yin Huang,
Xiaoxiao Li
and
Hamid Jafarkhani
\thanks{This work was supported in part by the NSF Award ECCS-2207457.
Tomas Ortega and Hamid Jafarkhani are with the University of California, Irvine, CA, USA 92697 (e-mail: \{tomaso, hamidj\}@uci.edu).
Chun-Yin Huang and Xiaoxiao Li are with the Computer Engineering Department at the University of British Columbia, Vancouver, BC, Canada, and Vector Institute, Toronto, ON, Canada (e-mail: \{chunyinh, xiaoxiao.li\}@ece.ubc.ca).}
}

\begin{document}
\maketitle

\begin{abstract}
    Distributed learning, particularly Federated Learning (FL), faces a significant bottleneck in the communication cost, particularly the uplink transmission of client-to-server updates, which is often constrained by asymmetric bandwidth limits at the edge.
    Biased compression techniques are effective in practice, but require error feedback mechanisms to provide theoretical guarantees and to ensure convergence when compression is aggressive.
    Standard error feedback, however, relies on client-specific control variates, which violates user privacy and is incompatible with stateless clients common in large-scale FL\@.
    This paper proposes two novel frameworks that enable biased compression without client-side state or control variates.
    The first, \cafelong{}, uses the globally aggregated update from the previous round as a shared control variate for all clients.
    The second, \cafeslong{}, extends this idea to scenarios where the server possesses a small private dataset; it generates a server-guided candidate update to be used as a more accurate predictor.
    We consider \dgdlong{} as a representative algorithm and analytically prove \cafe{}'s superiority to \dcgdlong{} with biased compression in the non-convex regime with bounded gradient dissimilarity.
    We further prove that \cafes{} converges to a stationary point, with a rate that improves as the server's data become more representative.
    Experimental results in FL scenarios validate the superiority of our approaches over existing compression schemes.
\end{abstract}


\section{Introduction}\label{sec-intro}
In distributed learning, a central server coordinates the training of a global model using data distributed across multiple clients.
The objective is to minimize a global loss function $f$, which is the average of the local loss functions of the clients
\begin{align}
    f(x) = \frac{1}{N} \sum_{n=1}^N f_n(x),\label{eq-global_loss}
\end{align}
where $x \in \R^d$ is the global model, $f_n \colon \R^d \to \R$ is the loss function of Client $n$, and $N$ is the total number of clients.
This formulation is prevalent in Federated Learning (FL)~\cite{communication_efficient}, a distributed learning paradigm designed for privacy preservation, where clients train the global model on their local data and send updates to the server for aggregation.

A primary bottleneck in distributed learning is communication overhead.
Given that communication at edge devices is not symmetric, with the upload (client to server) link being significantly more constrained~\cite{speedtest}, the download cost is considered negligible in the distributed learning literature.
Thus, the primary concern is the cost of uploading model updates from clients to the server~\cite{advances_open_problems}.
To mitigate this, various lossy compression techniques, such as quantization~\cite{qsgd}, sparsification~\cite{aji2017sparse}, and low-rank approximation~\cite{vogels2019powersgd}, have been proposed.
While unbiased compressors are easier to analyze, biased compressors are frequently preferred in practice for their superior empirical performance and computational efficiency~\cite{advances_open_problems,beznosikov2023biased}.

\begin{figure}
    \centering
    \tikzstyle{edge} = [very thick, -{Latex}]
\tikzstyle{vedge} = [very thick, magenta, dashdotted, -{Latex}]
\tikzstyle{my-circle} = [draw, circle, fill=white, very thick, minimum width=1cm, minimum height=1cm]
\tikzstyle{rounded-rectangle} = [draw, rectangle, rounded corners, fill=white, very thick, minimum width=2cm, minimum height=.75cm]
\tikzstyle{dashed-rounded-rectangle} = [draw, rectangle, rounded corners, fill=white, very thick, dashed, minimum width=2cm, minimum height=.75cm]

\begin{tikzpicture}

    \node[my-circle] (sum) {\textbf{$\sum$}};
    \node[xshift=-0.3cm, yshift=-0.2cm] (minus) at (sum.south) {\large{$-$}};
    \node[xshift=-0.2cm, yshift=0.3cm] (minus) at (sum.west) {\large{$+$}};
    \node[rounded-rectangle] (state) [left = of sum] {\textbf{Model}};
    \node[rounded-rectangle] (quantizer) [right = of sum] {\textbf{Compression}};
    \node[dashed-rounded-rectangle] (controlvar) [below = of quantizer] {\textbf{Control Variate}};

    \draw[edge] (state) -- (sum);
    \draw[edge] (sum) -- (quantizer);
    \draw[edge] (quantizer) -- (controlvar);
    \draw[edge] (controlvar) -| (sum);

\end{tikzpicture}
    \caption{Error feedback block diagram. Typically, each client stores its own control variate and must update it after each training round. The server stores its own copy of each client's control variate.}\label{fig:block-diagram}
\end{figure}
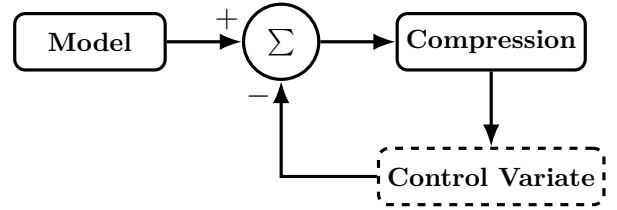

To guarantee the convergence of algorithms using biased compression, a mechanism known as error feedback (or error compensation) is typically required~\cite{error_feedback}.
This involves each client tracking the accumulated compression error locally and correcting for it in subsequent updates.
\Cref{fig:block-diagram} shows the block diagram for an error feedback mechanism.
The control variate shown in the figure is precisely the variable responsible for accumulating the compression error at each client.
However, standard error feedback requires both parties (clients and server) to store these error vectors in their respective control variates.
In other words, the clients and the server must maintain a state between rounds to be able to compensate for past compression errors.
This introduces two problems: (i) storing control variates at the clients and (ii) storing control variates at the server.
For the first issue, in many large-scale distributed learning scenarios, such as cross-device FL, clients are very unreliable and are thus continuously added to and dropped from the training pool throughout the rounds. Therefore, in most deployments, clients are assumed to be stateless and cannot maintain the state between communication rounds~\cite{advances_open_problems}.
For the latter issue, having the server store a control variate for each client introduces privacy risks by tracking individual client behavior over time, which completely defeats the purpose of many distributed training applications.

This paper addresses these challenges by proposing two novel frameworks that enable biased compression without requiring client-side state or per-client server-side tracking.
The core idea is to provide clients with a globally consistent predictor vector.
Clients then compress the \textbf{difference} between their local update and this predictor.
If the predictor is a good approximation of the local update, the difference vector will be sparse or will have a small magnitude, making it highly compressible.

We propose two distinct sources for this predictor, \cafelong{} and \cafeslong{}.
The first, \cafe{}, uses the aggregated and decoded update from the \emph{previous} communication round as a predictor for the current round $k$.
This approach is entirely self-contained, requires no additional data, and is suitable for any FL setting.
Clients receive this shared vector from the server along with the new model, use it for compression, and then discard it, thus remaining stateless.
The second framework, \cafes{}, is designed for scenarios where the server has access to a small private dataset relevant to the global learning task.
The server uses this dataset to compute a ``candidate update'' which serves as a more timely and potentially more accurate predictor than the lagged aggregate from \cafe.

Both frameworks are inspired by the use of temporal prediction in video coding, where compressing the difference between consecutive frames yields significant gains~\cite{993440}.
We analyze both methods in the \dgdlong{} setting for non-convex objectives.
Our theoretical contributions show that \cafe{} improves the convergence rate over standard compressed \dgd{}, while \cafes{} achieves a convergence rate that depends on the similarity between the server's data and the clients' data.
Experimental results on standard FL benchmarks confirm that our proposed methods outperform existing compression schemes.

To the best of our knowledge, this is the first work to propose a globally shared error feedback mechanism that eliminates the need for client-side memory.
Our main contributions are summarized as follows:
\begin{itemize}
    \item \textbf{Stateless Error Feedback:} We propose \cafe, a framework that leverages the aggregated update from the previous round as a global predictor. This allows clients to transmit compressed difference vectors without maintaining local error accumulation vectors, making it uniquely suitable for cross-device FL with stateless clients.
    \item \textbf{Server-Guided Optimization:} We introduce \cafes, which utilizes a server-side auxiliary dataset to generate candidate updates. This transforms the server from a passive aggregator into an active participant that guides the compression process, reducing uplink variance.
    \item \textbf{Theoretical Guarantees:} We provide a rigorous convergence analysis for both frameworks in the non-convex setting. We prove that \cafe{} achieves a convergence rate superior to standard \dcgd{} by a factor of $(1-\omega)$, and \cafes{} achieves a rate that improves linearly with the representativeness of the server's data.
\end{itemize}

The remainder of this paper is organized as follows.
We review related work in communication compression and error feedback in \Cref{sec-related_work}.
We then formally introduce the \cafe{} and \cafes{} frameworks in \Cref{sec-method}.
In \Cref{sec-analysis}, we provide a detailed convergence analysis for both methods, comparing their theoretical guarantees with the \dcgd{} baseline.
Finally, we present our experimental validation in \Cref{sec-experiments} and conclude in \Cref{sec-conclusion}.

\section{Related Work}\label{sec-related_work}
Communication compression is a well-studied topic in distributed learning, and error feedback is often suggested to improve convergence guarantees~\cite{SCAFFOLD}.
In~\cite{error_feedback}, the authors study the error feedback mechanism for one-bit per coordinate biased compression.
For general sparse compressors, error feedback was studied in~\cite{sparsifiedSGD,sparsifiedGradientMethods}.
For the decentralized setting,~\cite{chocosgd,Ortega_Jafarkhani_2023_gossiped} proposed variants of error feedback with general compression operators.
For asynchronous methods,~\cite{Ortega_Jafarkhani_2023_asynch,Ortega_Jafarkhani_2024a} also showed that a modified error feedback with general compression operators has good convergence guarantees.
In the non-convex setting,~\cite{EF21} showed that error feedback can be used in arbitrarily heterogeneous settings, which was later extended to the stochastic and convex settings in~\cite{beznosikov2023biased}.
All these methods, however, adhere to the traditional model of per-client error accumulation, making them unsuitable for stateless FL environments.
Our work departs from this by using a global, shared predictor, eliminating the need for client-specific states.

Table~\ref{tab:comparison} positions our work against key state-of-the-art baselines.
While MARINA~\cite{Gorbunov_Burlachenko_Li_Richtarik_2021} offers variance reduction for unbiased compression, and recent methods like EF21-P~\cite{Gruntkowska_Tyurin_Richtárik_2023} and ADEF~\cite{Gao_Rodomanov_Rack_Stich_2025} push the boundaries of convergence speed through bidirectional feedback and acceleration, they all share a common limitation: the requirement for \textit{client-side state}.
Whether storing error vectors, gradient estimators, or momentum buffers, these requirements hinder participation in stateless, cross-device Federated Learning.
Our frameworks, \cafe{} and \cafes, eliminate this dependency by using global, server-side predictors.

\begin{table*}[htbp]
    \centering
    \caption{Feature comparison with State-Of-The-Art distributed learning frameworks}\label{tab:comparison}
    \begin{threeparttable}
        \begin{tabular}{lcccc}
            \toprule
            \textbf{Algorithm}                                   & \textbf{Compression} & \textbf{Error Feedback}       & \textbf{Client State} & \textbf{Server Data} \\
            \midrule
            \textit{Standard Baselines}                          &                      &                               &                       &                      \\
            \dgd{}~\cite{communication_efficient}                & None                 & N/A                           & No                    & No                   \\
            \dcgd{}~\cite{qsgd}                                  & Possibly Biased      & None                          & No                    & No                   \\
            \midrule
            \textit{State-of-the-Art (Stateful)}                 &                      &                               &                       &                      \\
            EF21~\cite{EF21}                                     & Possibly Biased      & Local Control Variate         & \textbf{Yes}          & No                   \\
            MARINA~\cite{Gorbunov_Burlachenko_Li_Richtarik_2021} & Unbiased             & Gradient Estimator            & \textbf{Yes}          & No                   \\
            EF21-P~\cite{Gruntkowska_Tyurin_Richtárik_2023}      & Possibly Biased      & Bidirectional EF              & \textbf{Yes}          & No                   \\
            ADEF~\cite{Gao_Rodomanov_Rack_Stich_2025}            & Possibly Biased      & Accelerated Local EF          & \textbf{Yes}          & No                   \\
            \midrule 
            \textit{Proposed Frameworks (Stateless)}             &                      &                               &                       &                      \\
            \textbf{\cafe{} (Ours)}                              & Possibly Biased      & \textbf{Global Aggregated}    & \textbf{No}           & No                   \\
            \textbf{\cafes{} (Ours)}                             & Possibly Biased      & \textbf{Global Server-Guided} & \textbf{No}           & \textbf{Yes}         \\
            \bottomrule
        \end{tabular}
        \begin{tablenotes}
            \footnotesize
            \item \textbf{Client State:} Indicates if clients must persist a vector (e.g., error accumulator $e_n^k$, gradient estimator $h_n^k$, or momentum buffer) between rounds.
            \item \textit{ADEF} integrates Nesterov acceleration with error feedback, which requires maintaining local momentum states.
        \end{tablenotes}
    \end{threeparttable}
\end{table*}

\section{Proposed Frameworks}\label{sec-method}

To discuss the algorithm design, first, we must cover some compression preliminaries.
When clients send a message to the server, they first encode it using a function $E$.
The server decodes the received information using a function $D$.
We call these functions the encoder and decoder, respectively.
For a general compression mechanism, the composition $D(E(x)) \coloneqq \compress{x}$ is called a compression operator~\cite{sparsifiedSGD}.
\begin{definition}
    A compression operator is a function $\mathcal{C} \colon \R^d \to \R^d$, paired with a positive compression parameter $\omega < 1$, such that for any vector $x$,
    \begin{equation}\label{eq-compression}
        \ec{\sqn{\compress{x} - x}} \leq \omega \sqn{x}.
    \end{equation}
\end{definition}
\begin{example}[Top-k compression~\cite{qsgd}]\label{ex-topk}
    The top-k operator zeroes out all but the $k$ largest-magnitude elements of a vector.
    It satisfies \Cref{eq-compression} with $\omega = 1 - k/d$.
\end{example}

Next, we describe how compression operators are used when minimizing the global loss function from \Cref{eq-global_loss} in a distributed learning setting.
The fundamental algorithm for this purpose is \dcgdlong{} --- see \Cref{alg-classic-gd}.
The pseudocode shows how, at each round, the global model is sent to the clients, which train it using gradients computed with local data.
Clients then compress these gradients and send them to the server, which averages them to update the global model.
This process is repeated for any desired number of rounds.
\begin{algorithm}[htbp]
    \caption{Distributed Compressed Gradient Descent (\dcgd)}\label{alg-classic-gd}
    \begin{algorithmic}[1]
        \State \textbf{Input:} Initial model $x^0$,  Rounds $K$, Encoder-Decoder $(E,D)$ pair for compression, learning rate $\gamma$
        \State Initialize global model $x^0$, and aggregate $\Delta_s^0 \gets 0$
        \For{round $k$ from 0 to $K-1$}
        \State Send $x^k$ to all clients
        \For{each client $n$ in parallel}
        \State $y_n^k \gets x^k - \gamma \nabla f_n(x^k)$ \Comment{Train $x^k$ using local data, store the output in $y_n^k$}
        \State $\Delta_n^k \gets y_n^k - x^k =  - \gamma \nabla f_n(x^k)$ \Comment{Compute local update}
        \State Send $E(\Delta_n^k)$ to server \Comment{Upload local update}
        \EndFor
        \State Server decodes each client $n$ via $q_n^k \gets D(E(\Delta_n^k))$
        \State Server aggregates client updates $\Delta_s^k \coloneqq \frac{1}{N} \sum q_n^k$
        \State Server updates model $x^{k+1} \coloneqq x^k + \Delta_s^k$.
        \EndFor
    \end{algorithmic}
\end{algorithm}
Note that \dcgd{} is a specific instance of the general distributed learning framework, where we have chosen \gdlong{} as the optimizer for local models, and equal-weight averaging for the aggregation strategy.
We can derive a general strategy by not determining the aggregation strategy for client updates, nor the optimizer for on-client training.

\subsection{\cafe{} overview}
Our framework, \cafe, leverages the previous aggregated update $\Delta_s^{k-1}$ to help clients compute a more compressible update.
Namely, clients compress the difference between their local update $\Delta_n^k$ and the previous aggregated update
\begin{align*}
    E(\Delta_n^k - \Delta_s^{k-1}).
\end{align*}
On the server side, the server will add the previous aggregated update when decoding the received messages
\begin{align}
    q_n^k \gets D(E(\Delta_n^k - \Delta_s^{k-1})) + \Delta_s^{k-1}.\label{eq-cafe-decoding}
\end{align}
The pseudocode for this procedure is in \Cref{alg-cafe}, where the novelty with respect to the general distributed learning framework is highlighted in green boxes.
\begin{algorithm}[htbp]
    \caption{\cafelong{}}\label{alg-cafe}
    \begin{algorithmic}[1]
        \State \textbf{Input:} Initial model $x^0$, Rounds $K$, Encoder-Decoder $(E,D)$ pair for compression, learning rate $\gamma$
        \State Initialize global model $x^0$, and aggregate $\Delta_s^{-1} \gets 0$
        \For{round $k$ from 0 to $K-1$}
        \State Send $x^k$ \mybox{and $\Delta_s^{k-1}$} to all clients \Comment{In the stateful version $\Delta_s^{k-1}$ may be omitted}
        \For{each client $n$ in parallel}
        \State Compute local update $\Delta_n^k \gets - \gamma \nabla f_n(x^k)$
        \State Upload difference: send \mybox{$E(\Delta_n^k - \Delta_s^{k-1})$} to server
        \EndFor
        \State Server decodes each client $n$ via \newline
        \phantom{hacky fix} \mybox{$q_n^k \gets D(E(\Delta_n^k - \Delta_s^{k-1})) + \Delta_s^{k-1}$}
        \State Server aggregates $\Delta_s^k \coloneqq \frac{1}{N} \sum_{n=1}^N q_n^k$
        \State Server updates model $x^{k+1} \coloneqq x^k + \Delta_s^k$
        \EndFor
    \end{algorithmic}
\end{algorithm}

The intuition is that if the client gradients do not change drastically between rounds, $\Delta_n^k$ will be close to the average update $\Delta_s^{k-1}$, making their difference highly compressible.

This method allows for stateless participation, as the predictor $\Delta_s^{k-1}$ is provided by the server.
However, if a client participated in the previous round and retained the model $x^{k-1}$, they can locally recover the predictor via $\Delta_s^{k-1} = x^k - x^{k-1}$ without additional communication.
Explicit transmission of the predictor is therefore only strictly necessary for clients that are dropping in, returning after absence, or strictly memory-constrained.

In many popular distributed learning algorithms, $x^k$ and $x^{k-1}$ determine $\Delta_s^{k-1}$, like Distributed Gradient Descent, FedAvg, etc.
In other words, since the predictor $\Delta_s^{k-1}$ represents the net global model update, this framework naturally accommodates server-side momentum or adaptive optimizers (e.g., Adam).
In such cases, $\Delta_s^{k-1}$ captures the momentum-accumulated velocity, which stabilizes the update trajectory and often serves as a more accurate predictor of the next step than raw gradients alone.

Note that the error feedback mechanism in~\cite{EF21} is a special case of \cafe{} with a single client.
In this case, the aggregated update at the server is simply the client update, and we can analyze it as a control variate.
However, in the multi-client setting, the aggregated update is a combination of all client updates, which acts as a proxy for client-specific control variates and requires novel analysis, shown in \Cref{sec-analysis}.

\subsection{\cafes{} overview}
Our second proposal, \cafes, is designed for scenarios where the server holds a small private dataset, with a corresponding loss function $f_s(x)$.
Instead of using the lagged update $\Delta_s^{k-1}$, the server computes a candidate update, $\Delta_c^k = -\gamma \nabla f_s(x^k)$, using its own data and the current model $x^k$.
This candidate update is sent to clients as the predictor.
The procedure is detailed in \Cref{alg-cafe-s}.

\begin{algorithm}[htbp]
    \caption{\cafeslong{}}\label{alg-cafe-s}
    \begin{algorithmic}[1]
        \State \textbf{Input:} Initial model $x^0$, rounds $K$, compression $(E,D)$, learning rate $\gamma$, \myboxblue{server loss $f_s$}
        \For{round $k$ from 0 to $K-1$}
        \State \myboxblue{Server computes candidate $\Delta_c^k \gets - \gamma \nabla f_s(x^k)$}
        \State Server sends $x^k$ \myboxblue{and $\Delta_c^k$} to all clients
        \For{each client $n$ in parallel}
        \State Compute local update $\Delta_n^k \gets - \gamma \nabla f_n(x^k)$
        \State Upload difference: send \myboxblue{$E(\Delta_n^k - \Delta_c^k)$} to server
        \EndFor
        \State Server decodes each client $n$ via\newline
        \phantom{hacky fix} \myboxblue{$q_n^k \gets D(E(\Delta_n^k - \Delta_c^k)) + \Delta_c^k$}
        \State Server aggregates $\Delta_s^k \coloneqq \frac{1}{N} \sum_{n=1}^N q_n^k$
        \State Server updates model $x^{k+1} \coloneqq x^k + \Delta_s^k$
        \EndFor
    \end{algorithmic}
\end{algorithm}

If the server's data distribution is similar to the average distribution of clients, $\Delta_c^k$ will be a highly accurate predictor of $\Delta_n^k$, leading to greater compression gains.

\begin{remark}[Communication Trade-off]
    In the stateless configuration, the server must transmit the predictor $\Delta_s^{k-1}$ (for \cafe) or $\Delta_c^k$ (for \cafes) to the clients alongside the global model $x^k$.
    While this effectively doubles the downlink payload compared to standard \dgd, the trade-off is favorable in Federated Learning for several reasons.
    First, edge client connections (e.g., LTE/5G, residential broadband) are typically highly asymmetric, with downlink bandwidth significantly exceeding uplink bandwidth~\cite{speedtest}.
    Second, the predictor is a single global vector broadcast to all participants, whereas client updates are unique unicast transmissions.
    Therefore, maximizing the compression of the scarce unicast uplink capacity at the expense of the abundant broadcast downlink capacity is potentially a net-positive strategy.
\end{remark}

\section{Analysis}\label{sec-analysis}
Having formally defined both frameworks, we now proceed to analyze their convergence properties.
We choose \gd{} as the optimizer for local training, as it is representative of a wide range of distributed learning algorithms~\cite{advances_open_problems}.
We proceed with the following standard assumptions~\cite{advances_open_problems,SCAFFOLD}:
\begin{assumption}[L-Smoothness]\label{as-L-smooth}
    The objective function $f$ is $L$-smooth, which implies that it is differentiable, $\nabla f$ is $L$-Lipschitz, and
    \begin{equation}
        f(y) \leq f(x) + \ev{\nabla f(x), y - x} + \frac{L}{2} \sqn{y - x}.
    \end{equation}
    Also, the objective function $f$ is lower-bounded by $f^\star$.
\end{assumption}
\begin{assumption}[Bounded Local Dissimilarity]\label{as-bounded-dissimilarity}
    The local gradients have bounded dissimilarity, that is, there exists a $B^2 \geq 1$ such that
    \begin{align}
        \frac{1}{N} \sum_{n=1}^N \sqn{\nabla f_n (x)} \leq B^2 \sqn{\nabla f(x)}.
    \end{align}
\end{assumption}

For reference, we first state the convergence of the baseline \dcgd{} algorithm.
\begin{restatable}{theorem}{thmDcgd}[\dcgd{} Convergence]\label{thm-theorem-dcgd}
    Under \Cref{as-bounded-dissimilarity,as-L-smooth}, with a learning rate $\gamma \leq \frac{1}{L}$ and a compression parameter $\omega$ such that $\omega B^2 < 1$, \dcgd{} satisfies
    \begin{align}
        \frac{1}{K} \sum_{k=0}^{K-1} \ec{\sqn{\nabla f(x^k)}} & \leq \frac{2 (f(x^0) - f^\star)}{\gamma K} \cdot \frac{1}{1-\omega B^2}.
    \end{align}
\end{restatable}
\begin{proof}[Proof Sketch]
    The proof relies on a standard descent lemma~\cite[Section 1.2.3]{Nesterov_2018}.
    The main challenge is to bound the variance introduced by compression, $\ec{\sqn{\overline{e}^k}}$, which represents the average compression error.
    Using \Cref{as-bounded-dissimilarity}, we can show this error is bounded by the gradient norm: $\ec{\sqn{\overline{e}^k}} \leq \omega B^2 \ec{\sqn{\nabla f(x^k)}}$.
    Plugging this bound into the descent lemma and telescoping the sum over $K$ iterations yields the final result.
    The full proof with details is in \Cref{ap-proofs}
\end{proof}

\subsection{Analysis of CAFe}
The key benefit of \cafe{} is that it improves the dependency on the compression parameter $\omega$.
\begin{restatable}{theorem}{thmCAFe}[\cafe{} + \dgd{} Convergence]\label{thm-theorem-cafe}
    Under \Cref{as-bounded-dissimilarity,as-L-smooth}, with a learning rate
    \begin{equation}
        \gamma \leq \frac{1-\omega}{L(1+\omega)}\label{eq-gamma-condition}
    \end{equation}
    and $\omega B^2 < 1$, \cafe{} with \dgd{} satisfies
    \begin{align}
        \frac{1}{K} \sum_{k=0}^{K-1} \ec{\sqn{\nabla f(x^k)}} & \leq \frac{2 (f(x^0) - f^\star)}{\gamma K} \cdot \frac{1-\omega}{1-\omega B^2}.
    \end{align}
\end{restatable}
\begin{proof}[Proof Sketch]
    The analysis for \cafe{} is more involved because the predictor $\Delta_s^{k-1}$ creates a temporal dependency: the compression error at Round $k$ is linked to the aggregate update (and thus the error) from Round $k-1$.
    We cannot bound the error at Step $k$ purely by the state at Step $k$, as we do for \dcgd{}.

    The proof, detailed in \Cref{ap-proofs}, proceeds in two main parts.
    We first derive a bound that links the expected error at Step $k+1$, $\ec{\sqn{\overline{e}^{k+1}}}$, to the error at Step $k$, $\ec{\sqn{\overline{e}^{k}}}$ and the model gradients.
    Second, we define a Lyapunov function $\Psi^{k} \coloneqq \ec{f(x^{k}) + C \sqn{\overline{e}^{k}}}$ for a carefully chosen constant $C = \frac{\gamma}{2(1-\omega)}$.
    By combining our recursive error bound with the standard descent lemma, we show that this Lyapunov function decreases at each step ($\Psi^{k+1} \leq \Psi^k - (\text{progress term})$).
    Telescoping this inequality proves convergence.
\end{proof}

Comparing \Cref{thm-theorem-cafe} to \Cref{thm-theorem-dcgd}, \cafe{} introduces an improvement factor of $(1-\omega)$.
This is significant for aggressive compression where $\omega$ is close to 1.

\subsection{Analysis of CAFe-S}
The analysis for \cafes{} requires an additional assumption that quantifies the similarity between the server's data and the clients' data.
\begin{assumption}[Bounded Server-Client Dissimilarity]\label{as-server-client-dissimilarity}
    The local gradients and the server gradient have bounded dissimilarity, that is, there exists a $G^2 \geq 0$ such that
    \begin{align}
        \frac{1}{N} \sum_{n=1}^N \sqn{\nabla f_n(x) - \nabla f_s(x)} \leq G^2 \frac{1}{N} \sum_{n=1}^N \sqn{\nabla f_n(x)}.
    \end{align}
\end{assumption}
A small $G^2$ implies that the server's data is a good proxy for the aggregate client data.

\begin{restatable}{theorem}{thmCAFes}[\cafes{} + \dgd{} Convergence]\label{thm-theorem-cafe-s}
    Under \Cref{as-L-smooth,as-bounded-dissimilarity,as-server-client-dissimilarity}, with a learning rate $\gamma \leq \frac{1}{L}$, and a compression parameter $\omega$ such that $\omega G^2 B^2 < 1$, \cafes{} with \dgd{} satisfies
    \begin{align}
        \frac{1}{K} \sum_{k=0}^{K-1} \ec{\sqn{\nabla f(x^k)}} & \leq \frac{2 (f(x^0) - f^\star)}{\gamma K (1-\omega G^2 B^2)}.
    \end{align}
\end{restatable}
\begin{proof}[Proof Sketch]
    The proof for \cafes{} has a structure very similar to that of \dcgd{} (\Cref{thm-theorem-dcgd}).
    The key difference is in how we bound the compression error $\ec{\sqn{\overline{e}^k}}$.

    Instead of compressing the raw update $\Delta_n^k$, clients compress the difference $\Delta_n^k - \Delta_c^k$.
    We bound the error of this compression using \Cref{as-server-client-dissimilarity,as-bounded-dissimilarity}, which shows that $\ec{\sqn{\overline{e}^k}} \leq (\omega G^2 B^2) \ec{\sqn{\nabla f(x^k)}}$.
    This error bound has the exact same form as the \dcgd{} bound, but with the factor $\omega B^2$ replaced by the (potentially much smaller) factor $\omega G^2 B^2$.
    The rest of the proof is identical to that of \Cref{thm-theorem-dcgd}.
    The full proof is in \Cref{ap-proofs}.
\end{proof}

\Cref{thm-theorem-cafe-s} shows that \cafes{} achieves a convergence rate that depends on the similarity between the server's data and the clients' data.
This result is intuitive: if the server's data is a perfect match ($G^2=0$), the algorithm achieves the same convergence rate as \dcgd{}.
For all values of $G^2 < 1$, \cafes{} outperforms \dcgd{}, which reflects the advantage of having a representative dataset at the server.

\subsection{Discussion of the theoretical results}
Let us highlight the implications of our theoretical findings.
Comparing \cafe{} (\Cref{thm-theorem-cafe}) with the baseline \dcgd{} (\Cref{thm-theorem-dcgd}), we observe an improvement factor of $(1-\omega)$ in the convergence bound for the same learning rate, which becomes significant as compression becomes more aggressive (i.e., as $\omega$ approaches 1).
This gain comes at the cost of a stricter learning rate condition, $\gamma \leq \frac{1-\omega}{L(1+\omega)}$, which is common for error-feedback methods.
In practice, however, we usually chose learning rates much smaller than $\frac{1}{L}$, so this condition is not a significant limitation.

Comparing \cafes{} (\Cref{thm-theorem-cafe-s}) with \dcgd{} (\Cref{thm-theorem-dcgd}), we see a different kind of improvement.
The convergence rate of \cafes{} depends on $1-\omega G^2 B^2$ instead of $1-\omega B^2$.
The new term $G^2$, defined in \Cref{as-server-client-dissimilarity}, quantifies the dissimilarity between the server's gradient and the average of the clients' gradients.
If the server's data represent the global data distribution, $G^2$ will be small ($G^2 < 1$), resulting in a faster convergence rate compared to \dcgd{}.
In the ideal case where the server's data is a perfect proxy ($G^2 \to 0$), the negative impact of both compression ($\omega$) and client heterogeneity ($B^2$) on the convergence rate is eliminated.
Importantly, \cafes{} achieves this improvement without imposing additional constraints on the learning rate, keeping the standard \dcgd{} condition, $\gamma \leq \frac{1}{L}$.

In short, the main take-aways from this analysis are as follows:
\cafe{} is a \emph{data-free} method.
It provides a \emph{guaranteed} convergence improvement of $(1-\omega)$ over standard compression, making it desirable for aggressive compression, but it requires a smaller learning rate.
In contrast, \cafes{} is a \emph{data-dependent} method.
It provides a \emph{conditional} improvement if the server's data represent the population (low $G^2$).
It has no learning rate penalty, but its performance depends on the quality of the server-side dataset.

\section{Experimental Results}\label{sec-experiments}

We now empirically validate the theoretical findings of \Cref{sec-analysis}.
Our experiments are designed to: (i) demonstrate the core working principle that our difference-based compression is effective, (ii) compare the end-to-end performance of \cafe{} against baselines with various datasets and compression methods, and (iii) validate the \cafes{} hypothesis (\Cref{thm-theorem-cafe-s}) that convergence performance is directly tied to the quality of the server's data.
The code to reproduce all experiments is publicly available~\cite{code_repo}.

\subsection{Experimental Setup}
We conducted FL experiments with 10 clients on MNIST~\cite{lecun1998gradient}, EMNIST~\cite{cohen2017emnist}, CIFAR-10 and CIFAR-100~\cite{krizhevsky2009learning} datasets.
We used \texttt{CONV4} models for MNIST/EMNIST, a \texttt{Dense6} model for CIFAR-10, and a \texttt{ResNet-20} model for CIFAR-100, following the setup in~\cite{isik2024adaptive}.
We present results for both homogeneous and heterogeneous data cases, denoted \emph{iid} and \emph{non-iid}, respectively.
For the latter, we randomly sampled $40\%$ of the total classes for each client.
For each round of communication, which we refer to as a global training round, all clients participated in training.
Clients performed one local \gd{} training epoch with a batch size of 64 per global training round.
We varied the number of global training rounds for each experiment.
We ran each experiment with 3 random seeds and report the accuracy means and standard deviations.

The learning rates for each dataset and model combination were tuned by performing a sweep over a set of seven log-spaced candidate values, ranging from $10^{-3}$ to $10^0$.
We selected the rate that yielded the highest accuracy during the specified number of rounds.
This process was performed independently for each method (combination of task and compressor), and we found that the optimal rates were approximately consistent across both frameworks.
The exact learning rates used in each experiment are tabulated in \Cref{tab-lr-all}.
We also provide a detailed sensitivity analysis for the learning rate and further details on the compressor setup in \Cref{ap-experimental_details}.

\subsection{Validation of working principle}
The premise of our proposed frameworks is that the difference between the local update and a shared predictor is more compressible than the raw update itself.
To quantify this, we define the \textit{Compression Gain Ratio} for a predictor $P$ at Round $k$ as
\begin{equation}
    \frac{\norm{\Delta_n^k - P}}{\norm{\Delta_n^k}}.
\end{equation}
A ratio smaller than one indicates that the predictor reduces the magnitude of the vector to be compressed.

\begin{figure*}[htbp]
    \centering
    \includegraphics[width=16cm]{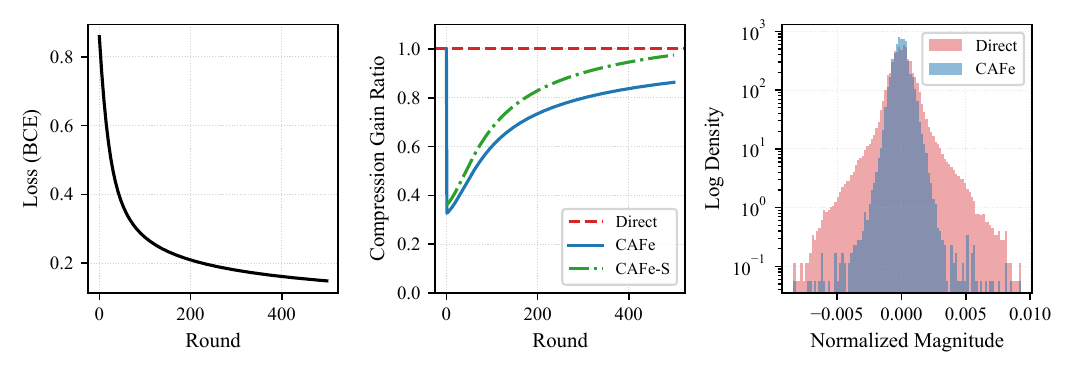}
    \vspace{-.3cm}
    \caption{Analysis of the \cafe{} working principle on a synthetic logistic regression task.
        \textbf{Left:} Global training loss showing convergence.
        \textbf{Middle:} The Compression Gain Ratio ($\rho$) remains below 1.0 for the majority of training, indicating that the difference vector has a smaller norm than the raw update. The ratio approaches 1.0 as the model converges.
        \textbf{Right:} The log-density histogram shows that \cafe{} updates are more peaked at zero (sparser) than direct updates.}\label{fig-compress-viz}
\end{figure*}

To visualize this principle, we simulated a Federated Logistic Regression task using synthetic data ($D=200$, 10 clients, iid data, and a server-side dataset of the same size) over 500 communication rounds, with no compression.
We log the updates that would be sent with no modification, with \cafe{}, and with \cafes{}.
\Cref{fig-compress-viz} presents the analysis of the training dynamics and the corresponding logs.
The left panel displays the global training loss, confirming that the model converges to the optimum.
The middle panel plots the Compression Gain Ratio.
During the active training phase, \cafe{} consistently achieves a ratio below $1.0$, confirming that the difference vector's magnitude is smaller than that of the raw gradient.
We observe that as the model converges and the gradients vanish (approaching Round 500), the ratio trends toward $1.0$.
This occurs because both the update $\Delta_n^k$ and the predictor $\Delta_s^{k-1}$ approach zero, and the relative benefit of the predictor diminishes as the updates become dominated by stochastic noise.
The right panel shows the log-density histogram of the update values aggregated across the entire training process.
The \cafe{} distribution (blue) is significantly narrower compared to the direct updates (red).
For clarity, we omit the \cafes{} distribution, which closely resembles that of \cafe{}.
This increased sparsity suggests that the difference vectors are more efficient to compress.

\subsection{Performance of \cafe{}}
\begin{figure*}[!htbp]
    \centering
    \includegraphics[width=16cm]{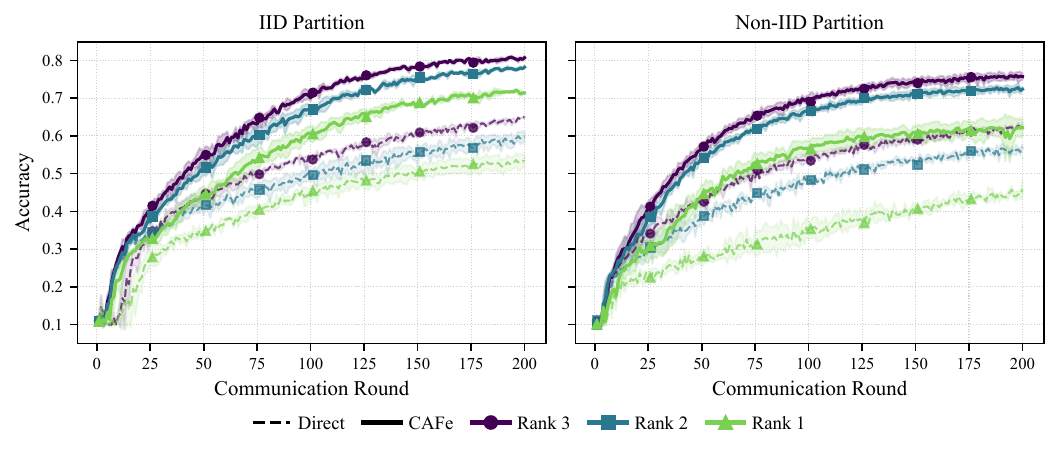}
    \vspace{-.3cm}
    \caption{Learning curves comparing direct compression (Direct) and \cafe{} with Low-rank compression in the CIFAR-10 task for iid and non-iid settings.~\cafe{} results are presented with solid lines, and Direct compression results are presented with dashed lines.~\cafe{} demonstrates faster convergence and higher accuracy.}\label{fig-perf}
\end{figure*}
\begin{table*}[!t]
    \centering
    \caption{Accuracy for \cafe{} and Direct compression (MNIST, EMNIST).}\label{tab-perf-mnist}
        \begin{tabular}{@{}clcccccccc@{}}
            \toprule
            & & \multicolumn{4}{c}{\textbf{MNIST} (\texttt{CONV4}, 50 rounds)} & \multicolumn{4}{c}{\textbf{EMNIST} (\texttt{CONV4}, 50 rounds)} \\
            \cmidrule(lr){3-6} \cmidrule(lr){7-10}
            & & \multicolumn{2}{c}{\textbf{iid}} & \multicolumn{2}{c}{\textbf{non-iid}} & \multicolumn{2}{c}{\textbf{iid}} & \multicolumn{2}{c}{\textbf{non-iid}} \\
            & & \multicolumn{2}{c}{$10$ classes/client} & \multicolumn{2}{c}{$4$ classes/client} & \multicolumn{2}{c}{$47$ classes/client} & \multicolumn{2}{c}{$4$ classes/client} \\
            \textbf{Compressor} & \textbf{Param.} & \textbf{Direct} & \textbf{\cafe} & \textbf{Direct} & \textbf{\cafe} & \textbf{Direct} & \textbf{\cafe} & \textbf{Direct} & \textbf{\cafe} \\
            \midrule
            None & --- & \multicolumn{2}{c}{$99.3 \pm 0.0$} & \multicolumn{2}{c}{$99.0 \pm 0.0$} & \multicolumn{2}{c}{$88.9 \pm 0.1$} & \multicolumn{2}{c}{$82.2 \pm 0.2$} \\
            \midrule
            \multirow{3}{*}{Top-k} & k=10\% & $\mathbf{99.4 \pm 0.0}$ & $99.4 \pm 0.0$ & $99.0 \pm 0.1$ & $\mathbf{99.0 \pm 0.1}$ & $\mathbf{88.9 \pm 0.0}$ & $88.9 \pm 0.1$ & $\mathbf{81.8 \pm 0.6}$ & $81.7 \pm 0.1$ \\
             & k=1\% & $99.2 \pm 0.0$ & $\mathbf{99.3 \pm 0.0}$ & $98.7 \pm 0.1$ & $\mathbf{98.9 \pm 0.1}$ & $88.0 \pm 0.1$ & $\mathbf{88.3 \pm 0.1}$ & $79.5 \pm 0.2$ & $\mathbf{81.4 \pm 0.3}$ \\
             & k=0.1\% & $98.7 \pm 0.1$ & $\mathbf{99.0 \pm 0.1}$ & $97.6 \pm 0.3$ & $\mathbf{98.1 \pm 0.1}$ & $85.6 \pm 0.3$ & $\mathbf{86.9 \pm 0.2}$ & $69.0 \pm 2.3$ & $\mathbf{75.2 \pm 0.2}$ \\
            \midrule
            \multirow{3}{*}{\begin{tabular}{@{}c@{}}Top-1\% +\\ Quantizer\end{tabular}} & 6 bits & $\mathbf{99.3 \pm 0.0}$ & $99.2 \pm 0.1$ & $98.7 \pm 0.1$ & $\mathbf{98.9 \pm 0.1}$ & $88.1 \pm 0.1$ & $\mathbf{88.3 \pm 0.1}$ & $79.7 \pm 0.7$ & $\mathbf{81.3 \pm 0.2}$ \\
             & 5 bits & $\mathbf{99.3 \pm 0.0}$ & $99.3 \pm 0.0$ & $98.6 \pm 0.1$ & $\mathbf{98.9 \pm 0.1}$ & $88.2 \pm 0.1$ & $\mathbf{88.5 \pm 0.1}$ & $80.6 \pm 0.2$ & $\mathbf{81.7 \pm 0.8}$ \\
             & 4 bits & $\mathbf{99.3 \pm 0.0}$ & $99.2 \pm 0.1$ & $98.5 \pm 0.1$ & $\mathbf{98.8 \pm 0.1}$ & $88.0 \pm 0.2$ & $\mathbf{88.4 \pm 0.1}$ & $79.1 \pm 0.6$ & $\mathbf{81.2 \pm 0.3}$ \\
            \midrule
            \multirow{3}{*}{Low-rank} & rank=3 & $99.0 \pm 0.1$ & $\mathbf{99.2 \pm 0.1}$ & $98.4 \pm 0.1$ & $\mathbf{98.8 \pm 0.0}$ & $86.5 \pm 0.1$ & $\mathbf{88.3 \pm 0.1}$ & $74.6 \pm 1.1$ & $\mathbf{79.7 \pm 0.6}$ \\
             & rank=2 & $98.9 \pm 0.1$ & $\mathbf{99.2 \pm 0.1}$ & $98.2 \pm 0.1$ & $\mathbf{98.8 \pm 0.0}$ & $85.8 \pm 0.3$ & $\mathbf{87.9 \pm 0.0}$ & $67.3 \pm 2.5$ & $\mathbf{76.7 \pm 0.2}$ \\
             & rank=1 & $98.5 \pm 0.0$ & $\mathbf{99.2 \pm 0.1}$ & $97.4 \pm 0.1$ & $\mathbf{98.5 \pm 0.2}$ & $83.5 \pm 0.2$ & $\mathbf{87.0 \pm 0.1}$ & $50.0 \pm 3.2$ & $\mathbf{71.0 \pm 0.4}$ \\
            \midrule
            \multirow{3}{*}{\begin{tabular}{@{}c@{}}Rank-1 +\\ Quantizer\end{tabular}} & 5 bits & $98.5 \pm 0.0$ & $\mathbf{99.2 \pm 0.0}$ & $97.2 \pm 0.4$ & $\mathbf{98.5 \pm 0.2}$ & $83.4 \pm 0.5$ & $\mathbf{87.0 \pm 0.1}$ & $49.6 \pm 3.4$ & $\mathbf{71.6 \pm 0.9}$ \\
             & 4 bits & $98.6 \pm 0.1$ & $\mathbf{99.2 \pm 0.0}$ & $97.4 \pm 0.1$ & $\mathbf{98.6 \pm 0.0}$ & $83.6 \pm 0.3$ & $\mathbf{87.0 \pm 0.0}$ & $48.1 \pm 3.4$ & $\mathbf{71.3 \pm 0.6}$ \\
             & 3 bits & $98.6 \pm 0.1$ & $\mathbf{99.2 \pm 0.0}$ & $97.3 \pm 0.2$ & $\mathbf{98.6 \pm 0.0}$ & $83.7 \pm 0.2$ & $\mathbf{87.0 \pm 0.2}$ & $44.7 \pm 3.1$ & $\mathbf{70.3 \pm 0.7}$ \\
            \bottomrule
        \end{tabular}%
\end{table*}
\begin{table*}[!t]
    \centering
    \caption{Accuracy for \cafe{} and Direct compression (CIFAR-10, CIFAR-100).}\label{tab-perf-cifar}
        \begin{tabular}{@{}clcccccccc@{}}
            \toprule
            & & \multicolumn{4}{c}{\textbf{CIFAR-10} (\texttt{CONV6}, 200 rounds)} & \multicolumn{4}{c}{\textbf{CIFAR-100} (\texttt{ResNet-20}, 200 rounds)} \\
            \cmidrule(lr){3-6} \cmidrule(lr){7-10}
            & & \multicolumn{2}{c}{\textbf{iid}} & \multicolumn{2}{c}{\textbf{non-iid}} & \multicolumn{2}{c}{\textbf{iid}} & \multicolumn{2}{c}{\textbf{non-iid}} \\
            & & \multicolumn{2}{c}{$10$ classes/client} & \multicolumn{2}{c}{$4$ classes/client} & \multicolumn{2}{c}{$100$ classes/client} & \multicolumn{2}{c}{$40$ classes/client} \\
            \textbf{Compressor} & \textbf{Param.} & \textbf{Direct} & \textbf{\cafe} & \textbf{Direct} & \textbf{\cafe} & \textbf{Direct} & \textbf{\cafe} & \textbf{Direct} & \textbf{\cafe} \\
            \midrule
            None & --- & \multicolumn{2}{c}{$88.5 \pm 0.3$} & \multicolumn{2}{c}{$84.4 \pm 0.3$} & \multicolumn{2}{c}{$41.3 \pm 1.6$} & \multicolumn{2}{c}{$39.3 \pm 1.3$} \\
            \midrule
            \multirow{3}{*}{Top-k} & k=10\% & $86.3 \pm 0.2$ & $\mathbf{87.5 \pm 0.5}$ & $81.5 \pm 1.0$ & $\mathbf{83.9 \pm 0.6}$ & $33.1 \pm 0.8$ & $\mathbf{37.0 \pm 0.4}$ & $32.1 \pm 1.0$ & $\mathbf{34.4 \pm 2.3}$ \\
             & k=1\% & $74.1 \pm 0.7$ & $\mathbf{80.5 \pm 0.8}$ & $70.3 \pm 0.3$ & $\mathbf{76.8 \pm 0.6}$ & $19.8 \pm 0.8$ & $\mathbf{23.1 \pm 0.7}$ & $19.9 \pm 0.4$ & $\mathbf{21.7 \pm 2.0}$ \\
             & k=0.1\% & $52.9 \pm 1.6$ & $\mathbf{61.9 \pm 2.4}$ & $48.2 \pm 1.5$ & $\mathbf{60.8 \pm 2.8}$ & $12.1 \pm 0.7$ & $\mathbf{12.3 \pm 1.2}$ & $\mathbf{11.0 \pm 0.4}$ & $10.7 \pm 0.3$ \\
            \midrule
            \multirow{3}{*}{\begin{tabular}{@{}c@{}}Top-1\% +\\ Quantizer\end{tabular}} & 6 bits & $74.5 \pm 0.8$ & $\mathbf{77.5 \pm 1.5}$ & $69.4 \pm 1.4$ & $\mathbf{77.0 \pm 0.9}$ & $20.1 \pm 0.1$ & $\mathbf{23.4 \pm 1.2}$ & $20.0 \pm 0.6$ & $\mathbf{22.0 \pm 0.5}$ \\
             & 5 bits & $72.9 \pm 0.3$ & $\mathbf{75.3 \pm 0.6}$ & $69.7 \pm 2.1$ & $\mathbf{76.5 \pm 1.1}$ & $20.0 \pm 0.8$ & $\mathbf{24.0 \pm 0.5}$ & $20.7 \pm 0.7$ & $\mathbf{22.0 \pm 0.7}$ \\
             & 4 bits & $13.1 \pm 2.7$ & $\mathbf{72.5 \pm 2.0}$ & $69.2 \pm 0.3$ & $\mathbf{77.0 \pm 0.9}$ & $20.7 \pm 0.0$ & $\mathbf{24.3 \pm 0.2}$ & $19.1 \pm 0.7$ & $\mathbf{21.7 \pm 0.4}$ \\
            \midrule
            \multirow{3}{*}{Low-rank} & rank=3 & $65.0 \pm 0.4$ & $\mathbf{80.8 \pm 0.4}$ & $62.7 \pm 0.9$ & $\mathbf{76.0 \pm 0.8}$ & $19.3 \pm 0.3$ & $\mathbf{32.4 \pm 0.5}$ & $19.5 \pm 0.7$ & $\mathbf{26.6 \pm 0.6}$ \\
             & rank=2 & $59.9 \pm 1.2$ & $\mathbf{78.2 \pm 0.1}$ & $57.0 \pm 0.7$ & $\mathbf{72.8 \pm 1.1}$ & $16.5 \pm 1.2$ & $\mathbf{29.8 \pm 0.2}$ & $17.4 \pm 0.2$ & $\mathbf{24.6 \pm 1.0}$ \\
             & rank=1 & $53.6 \pm 1.9$ & $\mathbf{72.2 \pm 0.4}$ & $45.8 \pm 2.6$ & $\mathbf{62.2 \pm 2.4}$ & $12.9 \pm 0.5$ & $\mathbf{24.1 \pm 0.6}$ & $13.3 \pm 0.8$ & $\mathbf{17.6 \pm 1.2}$ \\
            \midrule
            \multirow{3}{*}{\begin{tabular}{@{}c@{}}Rank-1 +\\ Quantizer\end{tabular}} & 5 bits & $52.5 \pm 0.8$ & $\mathbf{71.8 \pm 0.4}$ & $43.6 \pm 1.4$ & $\mathbf{62.8 \pm 1.8}$ & $13.1 \pm 0.7$ & $\mathbf{24.3 \pm 0.5}$ & $12.2 \pm 0.3$ & $\mathbf{17.4 \pm 0.9}$ \\
             & 4 bits & $53.5 \pm 1.4$ & $\mathbf{70.7 \pm 0.5}$ & $46.2 \pm 2.3$ & $\mathbf{62.7 \pm 0.9}$ & $13.2 \pm 0.2$ & $\mathbf{23.5 \pm 0.7}$ & $14.0 \pm 0.5$ & $\mathbf{16.5 \pm 1.1}$ \\
             & 3 bits & $53.6 \pm 1.7$ & $\mathbf{68.4 \pm 0.7}$ & $44.1 \pm 1.5$ & $\mathbf{61.7 \pm 1.4}$ & $13.4 \pm 0.3$ & $\mathbf{22.6 \pm 0.8}$ & $13.6 \pm 1.1$ & $\mathbf{17.3 \pm 0.7}$ \\
            \bottomrule
        \end{tabular}%
\end{table*}
\begin{table*}[!t]
    \centering
    \caption{Communication cost (Bits Per Parameter) for \cafe{} and Direct compression (MNIST, EMNIST).}\label{tab-bpp-mnist}
        \begin{tabular}{@{}clcccccccc@{}}
            \toprule
            & & \multicolumn{4}{c}{\textbf{MNIST} (\texttt{CONV4}, 50 rounds)} & \multicolumn{4}{c}{\textbf{EMNIST} (\texttt{CONV4}, 50 rounds)} \\
            \cmidrule(lr){3-6} \cmidrule(lr){7-10}
            & & \multicolumn{2}{c}{\textbf{iid}} & \multicolumn{2}{c}{\textbf{non-iid}} & \multicolumn{2}{c}{\textbf{iid}} & \multicolumn{2}{c}{\textbf{non-iid}} \\
            & & \multicolumn{2}{c}{$10$ classes/client} & \multicolumn{2}{c}{$4$ classes/client} & \multicolumn{2}{c}{$47$ classes/client} & \multicolumn{2}{c}{$4$ classes/client} \\
            \textbf{Compressor} & \textbf{Param.} & \textbf{Direct} & \textbf{\cafe} & \textbf{Direct} & \textbf{\cafe} & \textbf{Direct} & \textbf{\cafe} & \textbf{Direct} & \textbf{\cafe} \\
            \midrule
            None & --- & \multicolumn{2}{c}{$3.20 \cdot 10^{1}$} & \multicolumn{2}{c}{$3.20 \cdot 10^{1}$} & \multicolumn{2}{c}{$3.20 \cdot 10^{1}$} & \multicolumn{2}{c}{$3.20 \cdot 10^{1}$} \\
            \midrule
            \multirow{3}{*}{Top-k} & k=10\% & $5.40 \cdot 10^{0}$ & $5.40 \cdot 10^{0}$ & $5.40 \cdot 10^{0}$ & $5.40 \cdot 10^{0}$ & $5.40 \cdot 10^{0}$ & $5.40 \cdot 10^{0}$ & $5.40 \cdot 10^{0}$ & $5.40 \cdot 10^{0}$ \\
             & k=1\% & $5.40 \cdot 10^{-1}$ & $5.40 \cdot 10^{-1}$ & $5.40 \cdot 10^{-1}$ & $5.40 \cdot 10^{-1}$ & $5.40 \cdot 10^{-1}$ & $5.40 \cdot 10^{-1}$ & $5.40 \cdot 10^{-1}$ & $5.40 \cdot 10^{-1}$ \\
             & k=0.1\% & $5.40 \cdot 10^{-2}$ & $5.40 \cdot 10^{-2}$ & $5.40 \cdot 10^{-2}$ & $5.40 \cdot 10^{-2}$ & $5.40 \cdot 10^{-2}$ & $5.40 \cdot 10^{-2}$ & $5.40 \cdot 10^{-2}$ & $5.40 \cdot 10^{-2}$ \\
            \midrule
            \multirow{3}{*}{\begin{tabular}{@{}c@{}}Top-1\% +\\ Quantizer\end{tabular}} & 6 bits & $2.80 \cdot 10^{-1}$ & $2.80 \cdot 10^{-1}$ & $2.77 \cdot 10^{-1}$ & $2.76 \cdot 10^{-1}$ & $2.80 \cdot 10^{-1}$ & $2.80 \cdot 10^{-1}$ & $2.76 \cdot 10^{-1}$ & $2.78 \cdot 10^{-1}$ \\
             & 5 bits & $2.66 \cdot 10^{-1}$ & $2.64 \cdot 10^{-1}$ & $2.55 \cdot 10^{-1}$ & $2.59 \cdot 10^{-1}$ & $2.68 \cdot 10^{-1}$ & $2.68 \cdot 10^{-1}$ & $2.46 \cdot 10^{-1}$ & $2.55 \cdot 10^{-1}$ \\
             & 4 bits & $1.96 \cdot 10^{-1}$ & $2.55 \cdot 10^{-1}$ & $1.38 \cdot 10^{-1}$ & $2.13 \cdot 10^{-1}$ & $2.47 \cdot 10^{-1}$ & $2.51 \cdot 10^{-1}$ & $1.57 \cdot 10^{-1}$ & $1.83 \cdot 10^{-1}$ \\
            \midrule
            \multirow{3}{*}{Low-rank} & rank=3 & $4.79 \cdot 10^{-1}$ & $4.79 \cdot 10^{-1}$ & $4.79 \cdot 10^{-1}$ & $4.79 \cdot 10^{-1}$ & $4.80 \cdot 10^{-1}$ & $4.80 \cdot 10^{-1}$ & $4.80 \cdot 10^{-1}$ & $4.80 \cdot 10^{-1}$ \\
             & rank=2 & $3.22 \cdot 10^{-1}$ & $3.22 \cdot 10^{-1}$ & $3.22 \cdot 10^{-1}$ & $3.22 \cdot 10^{-1}$ & $3.22 \cdot 10^{-1}$ & $3.22 \cdot 10^{-1}$ & $3.22 \cdot 10^{-1}$ & $3.22 \cdot 10^{-1}$ \\
             & rank=1 & $1.64 \cdot 10^{-1}$ & $1.64 \cdot 10^{-1}$ & $1.64 \cdot 10^{-1}$ & $1.64 \cdot 10^{-1}$ & $1.65 \cdot 10^{-1}$ & $1.65 \cdot 10^{-1}$ & $1.65 \cdot 10^{-1}$ & $1.65 \cdot 10^{-1}$ \\
            \midrule
            \multirow{3}{*}{\begin{tabular}{@{}c@{}}Rank-1 +\\ Quantizer\end{tabular}} & 5 bits & $2.59 \cdot 10^{-2}$ & $2.59 \cdot 10^{-2}$ & $2.59 \cdot 10^{-2}$ & $2.59 \cdot 10^{-2}$ & $2.60 \cdot 10^{-2}$ & $2.60 \cdot 10^{-2}$ & $2.60 \cdot 10^{-2}$ & $2.60 \cdot 10^{-2}$ \\
             & 4 bits & $2.08 \cdot 10^{-2}$ & $2.08 \cdot 10^{-2}$ & $2.08 \cdot 10^{-2}$ & $2.08 \cdot 10^{-2}$ & $2.08 \cdot 10^{-2}$ & $2.08 \cdot 10^{-2}$ & $2.08 \cdot 10^{-2}$ & $2.08 \cdot 10^{-2}$ \\
             & 3 bits & $1.56 \cdot 10^{-2}$ & $1.56 \cdot 10^{-2}$ & $1.56 \cdot 10^{-2}$ & $1.56 \cdot 10^{-2}$ & $1.57 \cdot 10^{-2}$ & $1.57 \cdot 10^{-2}$ & $1.57 \cdot 10^{-2}$ & $1.57 \cdot 10^{-2}$ \\
            \bottomrule
        \end{tabular}%
\end{table*}
\begin{table*}[!t]
    \centering
    \caption{Communication cost (Bits Per Parameter) for \cafe{} and Direct compression (CIFAR-10, CIFAR-100).}\label{tab-bpp-cifar}
        \begin{tabular}{@{}clcccccccc@{}}
            \toprule
            & & \multicolumn{4}{c}{\textbf{CIFAR-10} (\texttt{CONV6}, 200 rounds)} & \multicolumn{4}{c}{\textbf{CIFAR-100} (\texttt{ResNet-20}, 200 rounds)} \\
            \cmidrule(lr){3-6} \cmidrule(lr){7-10}
            & & \multicolumn{2}{c}{\textbf{iid}} & \multicolumn{2}{c}{\textbf{non-iid}} & \multicolumn{2}{c}{\textbf{iid}} & \multicolumn{2}{c}{\textbf{non-iid}} \\
            & & \multicolumn{2}{c}{$10$ classes/client} & \multicolumn{2}{c}{$4$ classes/client} & \multicolumn{2}{c}{$100$ classes/client} & \multicolumn{2}{c}{$40$ classes/client} \\
            \textbf{Compressor} & \textbf{Param.} & \textbf{Direct} & \textbf{\cafe} & \textbf{Direct} & \textbf{\cafe} & \textbf{Direct} & \textbf{\cafe} & \textbf{Direct} & \textbf{\cafe} \\
            \midrule
            None & --- & \multicolumn{2}{c}{$3.20 \cdot 10^{1}$} & \multicolumn{2}{c}{$3.20 \cdot 10^{1}$} & \multicolumn{2}{c}{$3.20 \cdot 10^{1}$} & \multicolumn{2}{c}{$3.20 \cdot 10^{1}$} \\
            \midrule
            \multirow{3}{*}{Top-k} & k=10\% & $5.40 \cdot 10^{0}$ & $5.40 \cdot 10^{0}$ & $5.40 \cdot 10^{0}$ & $5.40 \cdot 10^{0}$ & $5.10 \cdot 10^{0}$ & $5.10 \cdot 10^{0}$ & $5.10 \cdot 10^{0}$ & $5.10 \cdot 10^{0}$ \\
             & k=1\% & $5.40 \cdot 10^{-1}$ & $5.40 \cdot 10^{-1}$ & $5.40 \cdot 10^{-1}$ & $5.40 \cdot 10^{-1}$ & $5.10 \cdot 10^{-1}$ & $5.10 \cdot 10^{-1}$ & $5.10 \cdot 10^{-1}$ & $5.10 \cdot 10^{-1}$ \\
             & k=0.1\% & $5.40 \cdot 10^{-2}$ & $5.40 \cdot 10^{-2}$ & $5.40 \cdot 10^{-2}$ & $5.40 \cdot 10^{-2}$ & $5.11 \cdot 10^{-2}$ & $5.11 \cdot 10^{-2}$ & $5.11 \cdot 10^{-2}$ & $5.11 \cdot 10^{-2}$ \\
            \midrule
            \multirow{3}{*}{\begin{tabular}{@{}c@{}}Top-1\% +\\ Quantizer\end{tabular}} & 6 bits & $2.59 \cdot 10^{-1}$ & $2.17 \cdot 10^{-1}$ & $2.71 \cdot 10^{-1}$ & $2.75 \cdot 10^{-1}$ & $2.50 \cdot 10^{-1}$ & $2.50 \cdot 10^{-1}$ & $2.49 \cdot 10^{-1}$ & $2.48 \cdot 10^{-1}$ \\
             & 5 bits & $2.40 \cdot 10^{-1}$ & $2.41 \cdot 10^{-1}$ & $2.30 \cdot 10^{-1}$ & $2.57 \cdot 10^{-1}$ & $2.38 \cdot 10^{-1}$ & $2.40 \cdot 10^{-1}$ & $1.91 \cdot 10^{-1}$ & $2.12 \cdot 10^{-1}$ \\
             & 4 bits & $6.60 \cdot 10^{-4}$ & $1.64 \cdot 10^{-1}$ & $1.88 \cdot 10^{-1}$ & $2.08 \cdot 10^{-1}$ & $2.21 \cdot 10^{-1}$ & $2.22 \cdot 10^{-1}$ & $1.07 \cdot 10^{-1}$ & $1.74 \cdot 10^{-1}$ \\
            \midrule
            \multirow{3}{*}{Low-rank} & rank=3 & $5.21 \cdot 10^{-1}$ & $5.21 \cdot 10^{-1}$ & $5.21 \cdot 10^{-1}$ & $5.21 \cdot 10^{-1}$ & $2.43 \cdot 10^{0}$ & $2.43 \cdot 10^{0}$ & $2.43 \cdot 10^{0}$ & $2.43 \cdot 10^{0}$ \\
             & rank=2 & $3.54 \cdot 10^{-1}$ & $3.54 \cdot 10^{-1}$ & $3.54 \cdot 10^{-1}$ & $3.54 \cdot 10^{-1}$ & $1.68 \cdot 10^{0}$ & $1.68 \cdot 10^{0}$ & $1.68 \cdot 10^{0}$ & $1.68 \cdot 10^{0}$ \\
             & rank=1 & $1.87 \cdot 10^{-1}$ & $1.87 \cdot 10^{-1}$ & $1.87 \cdot 10^{-1}$ & $1.87 \cdot 10^{-1}$ & $9.26 \cdot 10^{-1}$ & $9.26 \cdot 10^{-1}$ & $9.26 \cdot 10^{-1}$ & $9.26 \cdot 10^{-1}$ \\
            \midrule
            \multirow{3}{*}{\begin{tabular}{@{}c@{}}Rank-1 +\\ Quantizer\end{tabular}} & 5 bits & $2.96 \cdot 10^{-2}$ & $2.96 \cdot 10^{-2}$ & $2.96 \cdot 10^{-2}$ & $2.96 \cdot 10^{-2}$ & $1.54 \cdot 10^{-1}$ & $1.54 \cdot 10^{-1}$ & $1.54 \cdot 10^{-1}$ & $1.54 \cdot 10^{-1}$ \\
             & 4 bits & $2.38 \cdot 10^{-2}$ & $2.38 \cdot 10^{-2}$ & $2.38 \cdot 10^{-2}$ & $2.38 \cdot 10^{-2}$ & $1.25 \cdot 10^{-1}$ & $1.25 \cdot 10^{-1}$ & $1.25 \cdot 10^{-1}$ & $1.25 \cdot 10^{-1}$ \\
             & 3 bits & $1.79 \cdot 10^{-2}$ & $1.79 \cdot 10^{-2}$ & $1.79 \cdot 10^{-2}$ & $1.79 \cdot 10^{-2}$ & $9.59 \cdot 10^{-2}$ & $9.59 \cdot 10^{-2}$ & $9.59 \cdot 10^{-2}$ & $9.59 \cdot 10^{-2}$ \\
            \bottomrule
        \end{tabular}%
\end{table*}

Having validated the premise, we now evaluate the end-to-end performance of \cafe{} against direct compression in FL\@.
We test four biased compression methods.
First, we use \emph{Top-k} (see~\Cref{ex-topk}), which retains only the $k$ elements with the largest magnitudes.
Second, we use \emph{Low-rank approximation} via PowerSGD~\cite{vogels2019powersgd}, which compresses the update matrix by decomposing it into the product of two smaller matrices of rank $r$ (which we vary from 1 to 3), effectively projecting the update onto a lower-dimensional subspace.
We also evaluate variants where these compressed outputs are further quantized to lower bit-widths.
Sparsification is performed before quantization since it is optimal for FL~\cite{harma2024effective}.
We also include an uncompressed baseline as a reference.
The results for various compression parameter settings are reported in~\Cref{tab-perf-mnist,tab-perf-cifar}.
The communication cost in bits per parameter (bpp) for each compression method and parameter setting is reported in~\Cref{tab-bpp-mnist,tab-bpp-cifar}.

The results align with our theory: \cafe{} consistently outperforms or matches existing direct compression methods.
This is particularly stark in aggressive compression settings (low bpp).
For instance, in the CIFAR-10 iid task with Top-1\% + 4-bit Quantization, Direct compression collapses to random chance ($13.0\%$), whereas \cafe{} recovers a performance of $72.5\%$.
Similarly, with Rank-1 Low-rank compression on CIFAR-10 (non-iid), \cafe{} improves accuracy from $45.8\%$ to $62.2\%$.
We plot the learning curves using Low-rank compression in~\Cref{fig-perf} to provide a visual comparison of convergence speed and accuracy.

\subsection{Validation of \cafes{}}
\begin{figure}[htbp]
    \centering
    \includegraphics[width=\columnwidth]{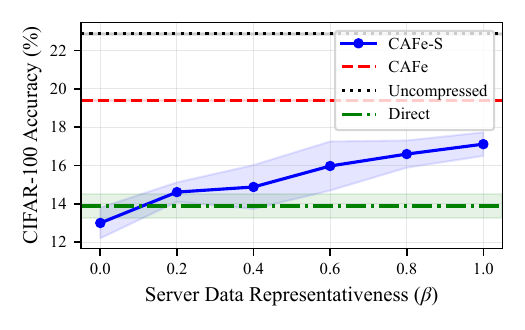}
    \caption{
        Accuracy of \cafes{} as a function of server's data representativeness ($\beta$) on CIFAR-100. While \cafes{} improves as the server's data become more representative, \cafe{} remains superior due to the stability of the aggregated predictor.}\label{fig-cafe-s-results}
\end{figure}
To validate our second proposed framework, \cafes{}, we designed an experiment to test the hypothesis of \Cref{thm-theorem-cafe-s}: that the convergence rate depends on the server-client dissimilarity, $G^2$.
We designed a sensitivity experiment using CIFAR-100.
Note that CIFAR-100 is an image classification dataset that has 100 classes grouped into 20 superclasses, making it suitable for creating related but distinct data distributions.

Clients received data from superclasses 0--9.
We then created a server-side dataset and tuned its ``representativeness'' using a parameter $\beta \in [0, 1]$.
For $\beta = 1.0$, the server's data were drawn entirely from superclasses 0--9 (the same as the clients), which makes the server a good proxy for the clients' data distribution ($G^2 \approx 0$).
For $\beta = 0.0$, the server's data were drawn entirely from superclasses 10--19, which makes the server a non-representative proxy.
For $\beta \in (0, 1)$, the server's data were drawn from both superclasses, creating a mixed proxy.
In all cases, the server's dataset size was fixed at $10\%$ of the total data.

We plot the accuracy of \cafes{} with rank-3 approximation compression against $\beta$ in \Cref{fig-cafe-s-results}, with horizontal baselines corresponding to the \cafe{} and uncompressed methods.
As predicted by our theory, the performance of \cafes{} improves monotonically as the server's data become more representative (increasing $\beta$).
However, we observe that \cafe{} outperforms \cafes{} even when $\beta=1.0$.
This reveals a trade-off between \emph{staleness} and \emph{variance}.
Although the \cafes{} predictor $\Delta_c^k$ is ``fresh'' (computed at Round $k$), it is generated from a small server split ($10\%$ of the total data), leading to higher variance.
In contrast, the \cafe{} predictor $\Delta_s^{k-1}$ is ``stale'' (from Round $k-1$), but is aggregated over all clients ($90\%$ of the data), making it a significantly more stable estimator of the global gradient direction.
This result suggests that while \cafes{} effectively leverages server's data, its advantage over the data-free \cafe{} is conditional on having a large enough dataset at the server to reduce variance below that of the aggregated history.

\section{Conclusion}\label{sec-conclusion}
We proposed two novel frameworks, \cafe{} and \cafes, to improve the efficiency of biased communication compression in distributed learning without relying on stateful clients or per-client error tracking.
\cafe{} leverages the previous global update as a shared predictor, making client updates more compressible.
\cafes{} extends this concept by using a candidate update generated from a server-side dataset, which can serve as a more accurate predictor.
We provided convergence guarantees for both methods in the non-convex setting, highlighting a trade-off between the data-free convergence improvement of \cafe{} and the potentially faster convergence of \cafes{}, which leverages a representative server-side dataset to mitigate the impact of inter-client heterogeneity.
Our experimental results for \cafe{} confirmed its practicality, showing significant improvements in convergence speed and model accuracy over standard compressed methods.

\appendices{}
\zcsetup{
    countertype = { section = appendix }
}

\section{Experimental Details}\label{ap-experimental_details}
Our experimental setup, including model architectures, best learning rate per scenario, and data distribution, is summarized in \Cref{tab-lr-all}.
\begin{table*}[!t]
    \centering
    \caption{Optimal learning rates for \cafe{} and Direct compression (MNIST, EMNIST, CIFAR-10, CIFAR-100).}\label{tab-lr-all}
    \resizebox{\textwidth}{!}{
        \begin{tabular}{@{}clcccccccccccccccc@{}}
            \toprule
            & & \multicolumn{4}{c}{\textbf{MNIST} (\texttt{CONV4}, 50 rounds)} & \multicolumn{4}{c}{\textbf{EMNIST} (\texttt{CONV4}, 50 rounds)} & \multicolumn{4}{c}{\textbf{CIFAR-10} (\texttt{CONV6}, 200 rounds)} & \multicolumn{4}{c}{\textbf{CIFAR-100} (\texttt{ResNet-20}, 200 rounds)} \\
            \cmidrule(lr){3-6} \cmidrule(lr){7-10} \cmidrule(lr){11-14} \cmidrule(lr){15-18}
            & & \multicolumn{2}{c}{\textbf{iid}} & \multicolumn{2}{c}{\textbf{non-iid}} & \multicolumn{2}{c}{\textbf{iid}} & \multicolumn{2}{c}{\textbf{non-iid}} & \multicolumn{2}{c}{\textbf{iid}} & \multicolumn{2}{c}{\textbf{non-iid}} & \multicolumn{2}{c}{\textbf{iid}} & \multicolumn{2}{c}{\textbf{non-iid}} \\
            & & \multicolumn{2}{c}{$10$ classes/client} & \multicolumn{2}{c}{$4$ classes/client} & \multicolumn{2}{c}{$47$ classes/client} & \multicolumn{2}{c}{$4$ classes/client} & \multicolumn{2}{c}{$10$ classes/client} & \multicolumn{2}{c}{$4$ classes/client} & \multicolumn{2}{c}{$100$ classes/client} & \multicolumn{2}{c}{$40$ classes/client} \\
            \textbf{Compressor} & \textbf{Param.} & \textbf{Direct} & \textbf{\cafe} & \textbf{Direct} & \textbf{\cafe} & \textbf{Direct} & \textbf{\cafe} & \textbf{Direct} & \textbf{\cafe} & \textbf{Direct} & \textbf{\cafe} & \textbf{Direct} & \textbf{\cafe} & \textbf{Direct} & \textbf{\cafe} & \textbf{Direct} & \textbf{\cafe} \\
            \midrule
            None & --- & \multicolumn{2}{c}{$10^{-0.5}$} & \multicolumn{2}{c}{$0.1$} & \multicolumn{2}{c}{$10^{-0.5}$} & \multicolumn{2}{c}{$0.1$} & \multicolumn{2}{c}{$10^{-0.5}$} & \multicolumn{2}{c}{$0.1$} & \multicolumn{2}{c}{$10^{-0.5}$} & \multicolumn{2}{c}{$1$} \\
            \midrule
            \multirow{3}{*}{Top-k} & k=10\% & $10^{-0.5}$ & $10^{-0.5}$ & $10^{-0.5}$ & $0.1$ & $10^{-0.5}$ & $10^{-0.5}$ & $0.1$ & $0.1$ & $10^{-0.5}$ & $0.1$ & $10^{-0.5}$ & $0.1$ & $10^{-0.5}$ & $10^{-0.5}$ & $1$ & $1$ \\
             & k=1\% & $10^{-0.5}$ & $10^{-0.5}$ & $0.1$ & $0.1$ & $10^{-0.5}$ & $10^{-0.5}$ & $0.1$ & $0.1$ & $10^{-0.5}$ & $0.1$ & $10^{-0.5}$ & $0.1$ & $1$ & $10^{-0.5}$ & $1$ & $1$ \\
             & k=0.1\% & $0.1$ & $0.1$ & $10^{-0.5}$ & $0.1$ & $0.1$ & $0.1$ & $0.1$ & $0.1$ & $10^{-0.5}$ & $0.1$ & $0.1$ & $0.1$ & $1$ & $0.1$ & $1$ & $10^{-0.5}$ \\
            \midrule
            \multirow{3}{*}{\begin{tabular}{@{}c@{}}Top-1\% +\\ Quantizer\end{tabular}} & 6 bits & $10^{-0.5}$ & $10^{-0.5}$ & $0.1$ & $0.1$ & $10^{-0.5}$ & $10^{-0.5}$ & $0.1$ & $0.1$ & $10^{-0.5}$ & $0.1$ & $10^{-0.5}$ & $0.1$ & $10^{-0.5}$ & $1$ & $1$ & $1$ \\
             & 5 bits & $10^{-0.5}$ & $10^{-0.5}$ & $0.1$ & $0.1$ & $10^{-0.5}$ & $10^{-0.5}$ & $10^{-0.5}$ & $0.1$ & $10^{-0.5}$ & $10^{-0.5}$ & $10^{-0.5}$ & $0.1$ & $1$ & $10^{-0.5}$ & $1$ & $1$ \\
             & 4 bits & $10^{-0.5}$ & $0.1$ & $0.1$ & $0.1$ & $10^{-0.5}$ & $10^{-0.5}$ & $0.1$ & $0.1$ & $0.01$ & $10^{-0.5}$ & $0.1$ & $0.1$ & $10^{-0.5}$ & $10^{-0.5}$ & $1$ & $10^{-0.5}$ \\
            \midrule
            \multirow{3}{*}{Low-rank} & rank=3 & $0.1$ & $10^{-0.5}$ & $0.1$ & $0.1$ & $0.1$ & $0.1$ & $0.1$ & $0.1$ & $10^{-0.5}$ & $0.1$ & $0.1$ & $0.1$ & $10^{-0.5}$ & $10^{-0.5}$ & $1$ & $10^{-0.5}$ \\
             & rank=2 & $0.1$ & $0.1$ & $0.1$ & $0.1$ & $10^{-0.5}$ & $0.1$ & $10^{-1.5}$ & $10^{-1.5}$ & $0.1$ & $0.1$ & $0.1$ & $0.1$ & $10^{-0.5}$ & $10^{-0.5}$ & $1$ & $10^{-0.5}$ \\
             & rank=1 & $0.1$ & $0.1$ & $0.1$ & $0.1$ & $0.1$ & $0.1$ & $10^{-1.5}$ & $10^{-1.5}$ & $10^{-0.5}$ & $0.1$ & $0.1$ & $0.1$ & $10^{-0.5}$ & $10^{-0.5}$ & $1$ & $10^{-0.5}$ \\
            \midrule
            \multirow{3}{*}{\begin{tabular}{@{}c@{}}Rank-1 +\\ Quantizer\end{tabular}} & 5 bits & $0.1$ & $0.1$ & $0.1$ & $0.1$ & $0.1$ & $0.1$ & $10^{-1.5}$ & $10^{-1.5}$ & $10^{-0.5}$ & $0.1$ & $0.1$ & $0.1$ & $10^{-0.5}$ & $10^{-0.5}$ & $1$ & $10^{-0.5}$ \\
             & 4 bits & $0.1$ & $0.1$ & $0.1$ & $0.1$ & $0.1$ & $0.1$ & $10^{-1.5}$ & $10^{-1.5}$ & $10^{-0.5}$ & $0.1$ & $10^{-0.5}$ & $0.1$ & $10^{-0.5}$ & $10^{-0.5}$ & $1$ & $10^{-0.5}$ \\
             & 3 bits & $0.1$ & $0.1$ & $0.1$ & $0.1$ & $0.1$ & $0.1$ & $10^{-1.5}$ & $10^{-1.5}$ & $0.1$ & $0.1$ & $0.1$ & $10^{-1.5}$ & $10^{-0.5}$ & $0.1$ & $1$ & $10^{-0.5}$ \\
            \bottomrule
        \end{tabular}%
    }
\end{table*}

\subsection{Compression Parameter Details}\label{ap-compression_params}
As reported in \Cref{tab-perf-mnist,tab-perf-cifar}, we select $k=10\%$, $1\%$, and $0.1\%$ for Top-k methods.
We also choose uniform quantization with 4, 5, and 6 bits for Top-k + Quantization and uniform quantization with 3, 4, and 5 bits for Low-rank approximation + Quantization.
For our Quantization experiments, we fix k to $1\%$ for Top-k and rank to $1$ for Low-rank approximation.
For Top-k + Quantization compression, we aim to test the lower limit of the choice for the number of bits per coordinate.
Notice in~\Cref{tab-perf-mnist,tab-perf-cifar} that when the number of bits is 4, Direct compression results in performance collapse on CIFAR-10, whereas \cafe{} maintains robustness.
For Low-rank compression, \cafe{} consistently outperforms direct compression by a large margin, regardless of the model architecture and dataset.
This is due to Low-rank approximation's low compression error, even when using rank 1.
With Low-rank approximation + Quantization, we also test the lower limit and find that \cafe{} allows for aggressive quantization (down to 3 bits) where Direct compression typically degrades.

\subsection{Learning Rate Sensitivity}\label{ap-learning_rate}
\begin{figure}[htbp]
    \centering
    \includegraphics{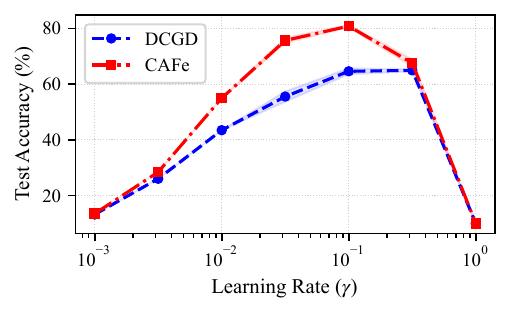}
    \caption{Learning rate sensitivity analysis on CIFAR-10 (iid) with rank-3 low-rank approximation compression.
    Both approaches achieve optimal performance around $\gamma=10^{-1}$ and $\gamma=10^{-0.5}$ and show similar shapes, suggesting the stricter theoretical bound for \cafe{} may not be a practical limitation.}\label{fig-ap-lr}
\end{figure}

\Cref{tab-lr-all} hints that both \cafe{} and direct compression tend to select similar learning rates.
In this section, we provide an ablation study on its sensitivity to the learning rate.
\Cref{thm-theorem-cafe} requires a stricter learning rate condition $\gamma \leq \frac{1-\omega}{L(1+\omega)}$ compared to \dcgd{}'s $\gamma \leq \frac{1}{L}$.
In \Cref{fig-ap-lr}, we plot the accuracy of both methods on CIFAR-10 (iid) as a function of the learning rate $\gamma$.
Using direct compression or \cafe{}, we find that both methods have similar performance curve shapes across a range of learning rates, albeit \cafe{}'s is always on top.
This suggests that the stricter theoretical bound, required for the Lyapunov proof, may not reflect the practical reality, and \cafe{} does not impose an additional hyperparameter tuning burden.

\section{Proofs of Convergence Theorems}\label{ap-proofs}
We now present the detailed proofs for the convergence theorems stated in \Cref{sec-analysis}, restated for clarity.
All three convergence proofs analyze the iterates of the form $x^{k+1} = x^k + \Delta_s^k$.
We first establish a common notation.
Let the local update be
\begin{equation}
    \Delta_n^k = - \gamma \nabla f_n(x^k)
\end{equation}
and the server's candidate update (for \cafes) be
\begin{equation}
    \Delta_c^k = - \gamma \nabla f_s(x^k).
\end{equation}
The compression error for a client $n$ at step $k$ is $e_n^k$, and the re-scaled error is $\hat e_n^k = e_n^k / \gamma$.
The average re-scaled error is $\overline{e}^k = \frac{1}{N} \sum_n \hat e_n^k$.
Then, the aggregate update $\Delta_s^k$ can always be written as:
\begin{align*}
    \Delta_s^k = \frac{1}{N}\sum_n (\Delta_n^k + e_n^k) = -\gamma \nabla f(x^k) + \gamma \overline{e}^k.
\end{align*}
This gives the unified iteration for all algorithms:
\begin{align}\label{eq-iteration-corrected}
    x^{k+1} = x^{k} + \Delta_s^k = x^{k} - \gamma \br{\nabla f (x^k) - \overline{e}^k}.
\end{align}
We define the effective gradient as $g^k \coloneqq \nabla f(x^k) - \overline{e}^k$, so the iteration is $x^{k+1} = x^k - \gamma g^k$.

All proofs start from the standard descent lemma, which follows from \Cref{as-L-smooth} and a learning rate $\gamma \leq \frac{1}{L}$:
\begin{align}
    f(x^{k+1}) & \leq f(x^k) + \ev{\nabla f(x^k), x^{k+1} - x^k} + \frac{L}{2} \sqn{x^{k+1} - x^k} \nonumber                          \\
               & = f(x^k) - \gamma \ev{\nabla f(x^k), g^k} + \frac{L\gamma^2}{2} \sqn{g^k} \nonumber                                  \\
               & \leq f(x^k) - \frac{\gamma}{2} \sqn{\nabla f(x^k)} - \br{\frac{\gamma}{2} - \frac{L\gamma^2}{2}} \sqn{g^k} \nonumber \\
               & \quad + \frac{\gamma}{2} \sqn{\nabla f(x^k) - g^k} \label{eq-descent-lemma-prev}                                     \\
               & \leq f(x^k) - \frac{\gamma}{2} \sqn{\nabla f(x^k)} + \frac{\gamma}{2} \sqn{\overline{e}^k}.\label{eq-descent-lemma}
\end{align}
The final inequality holds for $\gamma \leq \frac{1}{L}$ (which makes the $\sqn{g^k}$ term negative) and uses $\sqn{\nabla f(x^k) - g^k} = \sqn{\nabla f(x^k) - (\nabla f(x^k) - \overline{e}^k)} = \sqn{\overline{e}^k}$.
The proofs for the three theorems differ only in how they bound the $\ec{\sqn{\overline{e}^k}}$ term.

\subsection{Proof of Theorem~\ref{thm-theorem-dcgd} (\dcgd{})}
\thmDcgd*
\begin{proof}
    For standard \dcgd{} (\Cref{alg-classic-gd}), clients compress their raw local update $\Delta_n^k$.
    The error is $e_n^k = \compress{\Delta_n^k} - \Delta_n^k$.
    We bound the expectation of its squared norm using Jensen's inequality
    \begin{align*}
        \ec{\sqn{\overline{e}^k}} & = \ec{\sqn{\frac{1}{N} \sum_n \hat e_n^k}} \leq \frac{1}{N} \sum_n \ec{\sqn{\hat e_n^k}}  \\
                                  & = \frac{1}{N\gamma^2} \sum_n \ec{\sqn{e_n^k}}                                             \\
                                  & \leq \frac{\omega}{N\gamma^2} \sum_n \ec{\sqn{\Delta_n^k}} \tag{by \Cref{eq-compression}} \\
                                  & = \omega \frac{1}{N} \sum_n \ec{\sqn{\nabla f_n(x^k)}}                                    \\
                                  & \leq \omega B^2 \ec{\sqn{\nabla f(x^k)}}.\tag{by \Cref{as-bounded-dissimilarity}}
    \end{align*}
    Next, we substitute this error bound into our descent lemma (\Cref{eq-descent-lemma}) and take the expectation:
    \begin{align*}
        \ec{f(x^{k+1})} & \leq \ec{f(x^k) - \frac{\gamma}{2} \sqn{\nabla f(x^k)} + \frac{\gamma}{2} \sqn{\overline{e}^k}} \\
                        & = \ec{f(x^k)} - \frac{\gamma}{2} (1 - \omega B^2) \ec{\sqn{\nabla f(x^k)}}.
    \end{align*}
    Re-arranging and telescoping from $k=0$ to $K-1$ yields
    \begin{align*}
        \frac{\gamma}{2} (1 - \omega B^2) \sum_{k=0}^{K-1} \ec{\sqn{\nabla f(x^k)}} & \leq f(x^0) - \ec{f(x^K)} \\
                                                                                    & \leq f(x^0) - f^\star.
    \end{align*}
    Dividing by $K$ and re-arranging finishes the proof.
\end{proof}

\subsection{Proof of Theorem~\ref{thm-theorem-cafe} (\cafe{})}
The proof for \cafe{} is more complex as the error $\overline{e}^k$ depends on the previous update $\Delta_s^{k-1}$, which itself contains $\overline{e}^{k-1}$.
This requires a Lyapunov analysis.
We first need three supporting lemmas.
\begin{lemma}\label{lem-inner-product}
    Given an $L$-smooth function $f$, and iterations of the form $x^{k+1} = x^k - \gamma g^k$, we have
    \begin{align}
        -\ev{\nabla f(x^{k+1}), g^k} \leq -\ev{\nabla f(x^{k}), g^k} + \gamma L \sqn{g^k}.
    \end{align}
\end{lemma}
\begin{proof}
    We have
    \begin{align*}
        -\ev{\nabla f(x^{k+1}), g^k} = & -\ev{\nabla f(x^{k}), g^k}                       \\
                                       & + \ev{\nabla f(x^{k}) - \nabla f(x^{k+1}), g^k},
    \end{align*}
    and this can be bounded by
    \begin{align*}
        \ev{\nabla f(x^{k+1}) - \nabla f(x^k), g^k} & \leq \norm{\nabla f(x^{k+1}) - \nabla f(x^k)} \norm{g^k} \\
                                                    & \leq \gamma L \sqn{g^k},
    \end{align*}
    where we have used the $L$-smoothness of $f$ in the last step.
    Re-arranging, we obtain the desired result.
\end{proof}
\begin{lemma}\label{lem-compression}
    Given \Cref{as-bounded-dissimilarity,as-L-smooth}, the compression error for \dgd{} + \cafe{} (with $g^k = \nabla f(x^k) - \overline{e}^k$) satisfies
    \begin{equation}
        \begin{aligned}
            \ec{\sqn{\overline{e}^{k+1}}} & \leq \omega \ec{B^2 \sqn{\nabla f(x^{k+1})} - \sqn{\nabla f(x^k)}}    \\
                                          & + \gamma 2\omega L \ec{\sqn{g^k}} + \omega \ec{\sqn{\overline{e}^k}}.
        \end{aligned}
    \end{equation}
\end{lemma}
\begin{proof}
    We start by bounding the next-step error:
    \begin{align*}
        \ec{\sqn{\overline{e}^{k+1}}} & \leq \frac{1}{N} \sum_n\ec{\sqn{\hat e_n^{k+1}}} \tag{by Jensen's} \\
                                      & = \frac{1}{N\gamma^2} \sum_n\ec{\sqn{e_n^{k+1}}}
    \end{align*}
    In \cafe, clients compress $(\Delta_n^{k+1} - \Delta_s^k)$.
    So, by \Cref{eq-compression}, the last is upper bounded by
    \begin{align*}
         & \frac{\omega}{N\gamma^2} \sum_n \ec{ \sqn{\Delta_n^{k+1} - \Delta_s^k}}                    \\
         & = \frac{\omega}{N\gamma^2} \sum_n \ec{ \sqn{-\gamma \nabla f_n (x^{k+1}) - (-\gamma g^k)}} \\
         & = \frac{\omega}{N} \sum_n \ec{ \sqn{\nabla f_n (x^{k+1}) - g^k}}.
    \end{align*}
    We add and subtract $\nabla f(x^{k+1})$ inside the norm, and expand.
    Note that the cross-terms $\ev{\nabla f_n - \nabla f, \nabla f - g^k}$ sum over $n$ is zero, so the last expression equals
    \begin{equation}\label{eq-two-terms-lem-compression}
        \begin{aligned}
            \omega \ec{ \frac{1}{N} \sum_n \sqn{\nabla f_n (x^{k+1}) - \nabla f(x^{k+1})} } \\
            + \omega \ec{\sqn{\nabla f(x^{k+1}) - g^k}}
        \end{aligned}
    \end{equation}
    Using \Cref{as-bounded-dissimilarity} we can bound the first term in \Cref{eq-two-terms-lem-compression} as
    \begin{align*}
        \frac{1}{N}\sum_n\sqn{\nabla f_n - \nabla f} & = \br{\frac{1}{N}\sum_n\sqn{\nabla f_n}} - \sqn{\nabla f} \\
                                                     & \leq (B^2-1)\sqn{\nabla f}.
    \end{align*}
    Thus, to finish we expand the last term in \Cref{eq-two-terms-lem-compression},
    \begin{align*}
        \sqn{\nabla f(x^{k+1}) - g^k} = & \sqn{\nabla f(x^{k+1})} + \sqn{g^k} \\
                                        & - 2\ev{\nabla f(x^{k+1}), g^k},
    \end{align*}
    and apply \Cref{lem-inner-product} to bound the inner product term.
    This gives us
    \begin{align*}
        \sqn{\nabla f(x^{k+1}) - g^k} & \leq \br{\sqn{\nabla f(x^{k+1})} - \sqn{\nabla f(x^k)}}           \\
                                      & + \br{\sqn{\nabla f(x^k)} - 2\ev{\nabla f(x^k), g^k} + \sqn{g^k}} \\
                                      & + 2\gamma L \sqn{g^k}
    \end{align*}
    The middle term is $\sqn{\nabla f(x^k) - g^k} = \sqn{\overline{e}^k}$, so plugging this back into our main inequality gives the lemma.
\end{proof}
\begin{lemma}\label{lem-bound-gradient}
    Let $f\colon \mathbb{R}^d \to \mathbb{R}$ be an $L$-smooth function with a lower bound $f^\star$.
    Then, for any $x \in \mathbb{R}^d$,
    \begin{align*}
        \sqn{\nabla f(x)} & \leq 2L \br{f(x) - f^\star}.
    \end{align*}
\end{lemma}
\begin{proof}
    By \Cref{as-L-smooth}, for any $y$, we have
    \begin{align*}
        f(y) & \leq f(x) + \ev{\nabla f(x), y - x} + \frac{L}{2} \sqn{y - x}.
    \end{align*}
    We choose $y = x - \frac{1}{L} \nabla f(x)$, and obtain
    \begin{align*}
        f\left(y\right) & \leq f(x) - \frac{1}{2L} \sqn{\nabla f(x)}.
    \end{align*}
    Since $f(y) \geq f^\star$, we re-arrange and obtain the result.
\end{proof}

\thmCAFe*
\begin{proof}
    First, we define a Lyapunov function to capture both the function value and the compression error.
    Let
    \begin{equation}
        \Psi^{k} \coloneqq \ec{f(x^{k}) + \frac{\gamma}{2\br{1-\omega}} \sqn{\overline{e}^{k}}}.
    \end{equation}
    Our goal is to show that $\Psi^{k+1} \leq \Psi^k - (\text{progress term})$.
    We will combine the descent lemma (\Cref{eq-descent-lemma}) and the compression error bound from \Cref{lem-compression}.
    We start with the expectation of the descent lemma (\Cref{eq-descent-lemma-prev})
    \begin{align*}
        \ec{f(x^{k+1})} \leq & \ec{f(x^k)} - \frac{\gamma}{2} \ec{\sqn{\nabla f(x^k)}}         \\
                             & + \frac{\gamma}{2} \ec{\sqn{\overline{e}^k}}                    \\
                             & - \br{\frac{1}{2\gamma} - \frac{L}{2}} \ec{\sqn{x^{k+1} - x^k}}
    \end{align*}
    and $\frac{\gamma}{2(1-\omega)} \cdot \ec{\text{\Cref{lem-compression}}}$, which yields
    \begin{align*}
        \frac{\gamma \ec{\sqn{\overline{e}^{k+1}}}}{2(1-\omega)} \leq & \frac{\gamma \omega \ec{\sqn{\overline{e}^k}}}{2(1-\omega)}                         \\
                                                                      & + \frac{\gamma\omega}{2(1-\omega)} \ec{B^2 \sqn{\nabla f^{k+1}} - \sqn{\nabla f^k}} \\
                                                                      & + \frac{\gamma^2 \omega L}{1-\omega} \ec{\sqn{g^k}}.
    \end{align*}
    Summing these two inequalities gives
    \begin{align*}
        \Psi^{k+1} \leq & \underbrace{\br{\ec{f(x^k)} + \frac{\gamma \omega \ec{\sqn{\overline{e}^k}}}{2(1-\omega)} + \frac{\gamma}{2} \ec{\sqn{\overline{e}^k}}}}_{= \Psi^k}                        \\
                        & + \underbrace{\br{-\frac{\gamma}{2} \ec{\sqn{\nabla f^k}} - \frac{\gamma\omega}{2(1-\omega)} \ec{\sqn{\nabla f^k}}}}_{= -\frac{\gamma}{2(1-\omega)} \ec{\sqn{\nabla f^k}}} \\
                        & - \underbrace{\br{\frac{1}{2\gamma} - \frac{L}{2} - \frac{\gamma \omega L}{1-\omega}} \ec{\sqn{x^{k+1} - x^k}}}_{ \leq 0 \text{ by \Cref{eq-gamma-condition}}}             \\
                        & + \frac{\gamma\omega B^2}{2(1-\omega)} \ec{ \sqn{\nabla f(x^{k+1})} }.
    \end{align*}
    The $\Psi^k$ term simplifies because $\frac{\gamma\omega}{2(1-\omega)} + \frac{\gamma}{2} = \frac{\gamma}{2(1-\omega)}$.
    The $\sqn{\nabla f^k}$ term simplifies to $-\frac{\gamma(1-\omega) + \gamma\omega}{2(1-\omega)} = -\frac{\gamma}{2(1-\omega)}$.
    The $\sqn{x^{k+1}-x^k}$ term is negative due to \Cref{eq-gamma-condition}, so we can drop it.
    Thus,
    \begin{align*}
        \Psi^{k+1} \leq \Psi^k & - \frac{\gamma}{2(1-\omega)} \ec{\sqn{\nabla f(x^k)}}                  \\
                               & + \frac{\gamma\omega B^2}{2(1-\omega)} \ec{ \sqn{\nabla f(x^{k+1})} }.
    \end{align*}
    Given that $\omega B^2 < 1$, this almost gives us our desired Lyapunov decrease.
    We have taken the liberty of calling $\Psi^k$ the Lyapunov function, but it does not strictly decrease due to the last term, which depends on the next iterate $x^{k+1}$, instead of $x^k$.
    We will handle the extra term by first telescoping the recursion from $k=0$ to $K-1$,
    \begin{align*}
        \Psi^K \leq \Psi^0 & - \frac{\gamma}{2(1-\omega)} \sum_{k=0}^{K-1} \ec{\sqn{\nabla f(x^k)}}                  \\
                           & + \frac{\gamma\omega B^2}{2(1-\omega)} \sum_{k=0}^{K-1} \ec{ \sqn{\nabla f(x^{k+1})} }.
    \end{align*}
    Given $\Psi^0 = f(x^0)$, since $\overline{e}^0 = 0$, and $\Psi^K \geq f(x^K)$, we obtain
    \begin{align*}
        f(x^K) \leq f(x^0) & - \frac{\gamma(1 - \omega B^2)}{2(1-\omega)} \sum_{k=0}^{K-1} \ec{\sqn{\nabla f(x^k)}} \\
                           & + \frac{\gamma\omega B^2}{2(1-\omega)} \ec{\sqn{\nabla f(x^K)}},
    \end{align*}
    where we have dropped a negative $\sqn{\nabla f(x^0)}$ term.
    Next, we use \Cref{lem-bound-gradient} to bound the $\ec{\sqn{\nabla f(x^K)}}$ term,
    \begin{multline*}
        \frac{\gamma(1 - \omega B^2)}{2(1-\omega)} \sum_{k=0}^{K-1} \ec{\sqn{\nabla f(x^k)}}  \leq \\
        {} f(x^0) + \frac{\gamma\omega B^2}{2(1-\omega)} 2L \br{f(x^K) - f^\star} - f(x^K).
    \end{multline*}
    To obtain the theorem statement, it suffices to show that
    \begin{equation}\label{eq-final-step-cafe}
        \frac{\gamma\omega B^2}{2(1-\omega)} 2L \br{f(x^K) - f^\star} - f(x^K) \leq - f^\star,
    \end{equation}
    which holds if
    \begin{equation}
        \frac{\gamma\omega B^2}{1-\omega} L < 1.
    \end{equation}
    Notice that this is equivalent to \Cref{eq-gamma-condition}, given that $\omega B^2 < 1$.
    Thus, we have completed the proof.
\end{proof}

\subsection{Proof of Theorem~\ref{thm-theorem-cafe-s} (\cafes{})}
\thmCAFes*
\begin{proof}
    This proof follows the same structure as the \dcgd{} proof.
    For \cafes{}, clients compress the difference $\Delta_n^k - \Delta_c^k$.
    The error is $e_n^k = \compress{\Delta_n^k - \Delta_c^k} - (\Delta_n^k - \Delta_c^k)$.
    \begin{align*}
        \ec{\sqn{\overline{e}^k}} & \leq \frac{1}{N} \sum_n \ec{\sqn{\hat e_n^k}} \tag{by Jensen's}                                              \\
                                  & = \frac{1}{N\gamma^2} \sum_n \ec{\sqn{e_n^k}}                                                                \\
                                  & \leq \frac{\omega}{N\gamma^2} \sum_n \ec{\sqn{\Delta_n^k - \Delta_c^k}} \tag{by \Cref{eq-compression}}       \\
                                  & = \omega \frac{1}{N} \sum_n \ec{\sqn{\nabla f_n(x^k) - \nabla f_s(x^k)}}                                     \\
                                  & \leq \omega G^2 \frac{1}{N} \sum_n \ec{\sqn{\nabla f_n(x^k)}} \tag{by \Cref{as-server-client-dissimilarity}} \\
                                  & \leq \omega G^2 B^2 \ec{\sqn{\nabla f(x^k)}}.\tag{by \Cref{as-bounded-dissimilarity}}
    \end{align*}
    Now, notice that this error bound has the \emph{exact same form} as the \dcgd{} error bound, with the constant $\omega B^2$ simply replaced by $\omega G^2 B^2$.
    The remainder of the proof (telescoping the sum and re-arranging) is identical to the proof for \Cref{thm-theorem-dcgd}.
\end{proof}

\bibliographystyle{IEEEtran}
\bibliography{references}
\end{document}